\documentclass{article} 
\usepackage{iclr2024_conference,times}

\usepackage{amsmath,amsfonts,bm}

\def\eqref#1{equation~\ref{#1}}

\def\1{\bm{1}}

\DeclareMathAlphabet{\mathsfit}{\encodingdefault}{\sfdefault}{m}{sl}
\SetMathAlphabet{\mathsfit}{bold}{\encodingdefault}{\sfdefault}{bx}{n}

\usepackage{hyperref}
\usepackage{url}
\usepackage{booktabs}       
\usepackage{amsfonts}       
\usepackage{nicefrac}       
\usepackage{microtype}      
\usepackage{xcolor}         
\usepackage{savesym}
\usepackage{graphicx}
\usepackage{amsmath}
\usepackage{amssymb}
\usepackage{booktabs}
\usepackage{multirow}
\usepackage{wrapfig}
\usepackage{xcolor}
\usepackage{bbm}
\usepackage{float}
\usepackage{lipsum}  
\usepackage{breqn}
\usepackage[normalem]{ulem}
\useunder{\uline}{\ul}{}
\usepackage{enumitem}
\usepackage{subfig}
\usepackage{color, colortbl}
\usepackage{booktabs}
\usepackage{enumitem}
\usepackage{xcolor}
\usepackage{pifont}
\newcommand{\cmark}{\ding{51}}%
\newcommand{\xmark}{\ding{55}}%
\setlist[itemize]{align=parleft,left=0pt..1em}
\usepackage[normalem]{ulem}

\title{FeatUp: A Model-Agnostic Framework \\for Features at Any Resolution}

\author{Stephanie Fu\footnotemark[1] \space \footnotemark[2] \\
UC Berkeley \\
\And
Mark Hamilton\footnotemark[1]\\
MIT, Microsoft \\
\And
Laura Brandt \\
MIT
\AND
Axel Feldmann \\
MIT
\And
Zhoutong Zhang \\
Adobe Research
\And
William T. Freeman \\
MIT, Google
}

\iclrfinalcopy 
\begin{document}

\maketitle

\begin{figure}[h]
    \centering
    \vspace{-.15in}
    \includegraphics[width=1.0\linewidth]{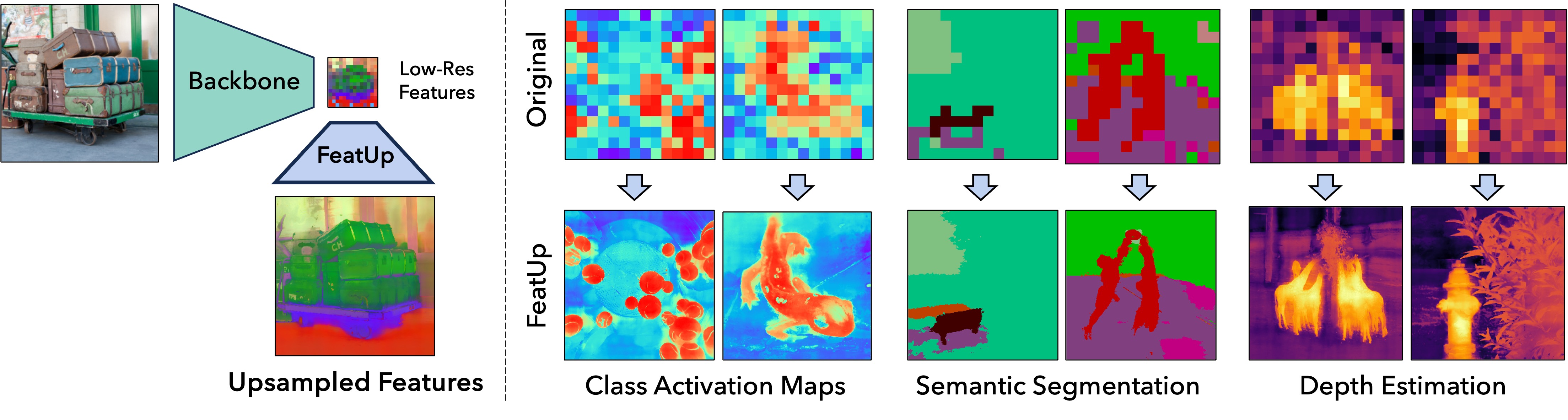}
    \vspace{-.22in}
    \caption{\small FeatUp upsamples image features from any model backbone, adding spatial resolution to existing semantics. High-res features can be learned either as a per-image implicit network or a general-purpose upsampling operation; the latter is a drop-in module to improve downstream dense prediction tasks.}
    \vspace{-.2in}

\end{figure}

  \setlength\extrarowheight{-3pt}
\let\thefootnote\relax\footnote{$^*$Equal contribution. Corresponding authors: \href{markth@mit.edu}{markth@mit.edu} and \href{fus@berkeley.edu}{fus@berkeley.edu}}
\let\thefootnote\relax\footnote{$^\dagger$Work done while at MIT.}
\begin{abstract}
Deep features are a cornerstone of computer vision research, capturing image semantics and enabling the community to solve downstream tasks even in the zero- or few-shot regime. However, these features often lack the spatial resolution to directly perform dense prediction tasks like segmentation and depth prediction because models aggressively pool information over large areas. In this work, we introduce FeatUp, a task- and model-agnostic framework to restore lost spatial information in deep features. We introduce two variants of FeatUp: one that guides features with high-resolution signal in a single forward pass, and one that fits an implicit model to a single image to reconstruct features at any resolution. Both approaches use a multi-view consistency loss with deep analogies to NeRFs. Our features retain their original semantics and can be swapped into existing applications to yield resolution and performance gains even without re-training. We show that FeatUp significantly outperforms other feature upsampling and image super-resolution approaches in class activation map generation, transfer learning for segmentation and depth prediction, and end-to-end training for semantic segmentation. 
\end{abstract}
\vspace{-.15in}

\section{Introduction}
\label{sec:intro}
\vspace{-0.1in}
Recently, considerable effort has been made to develop methods to extract features from data modalities such as vision \citep{hog,sift,weiss2016survey,moco,dino}, text \citep{mikolov2013efficient,devlin2018bert,gpt}, and audio \citep{wav2vec,hubert}. These features often form the backbone of different methods, including classification \citep{shao2014transfer}, weakly-supervised learning \citep{ahn2019weakly,STEGO}, semantic segmentation \citep{wang2020self}, optical flow \citep{liu2010sift,teed2020raft}, neural rendering \citep{kobayashi2022decomposing}, and more recently, image generation \citep{StableDiffusion}. Despite their immense success, deep features often sacrifice spatial resolution for semantic quality. For example, ResNet-50 \citep{rn50} produces $7\times7$ deep features from a $224 \times 224$ pixel input ($32\times$ resolution reduction). Even Vision Transformers (ViTs) \citep{vit} incur a significant resolution reduction, making it challenging to perform dense prediction tasks such as segmentation or depth estimation using these features alone. 

To mitigate these issues, we propose FeatUp: a novel framework to improve the resolution of any vision model's features without changing their original ``meaning'' or orientation. Our primary insight, inspired by 3D reconstruction frameworks like NeRF \citep{nerf}, is that multiview consistency of low-resolution signals can supervise the construction of high-resolution signals. More specifically, we learn high-resolution information by aggregating low resolution views from a model's outputs across multiple ``jittered'' (e.g. flipped, padded, cropped) images. We aggregate this information by learning an upsampling network with a multiview consistency loss. Our work explores two architectures for upsampling: a single guided upsampling feedforward network that generalizes across images, and an implicit representation overfit to a single image.

This feedforward upsampler is a parameterized generalization of a Joint Bilateral Upsampling (JBU) filter \citep{jbu} powered by a CUDA kernel orders of magnitude faster and less memory-intensive than existing implementations. This upsampler can produce high quality features aligned to object edges at a computational cost comparable to a few convolutions. Our implicit upsampler draws a direct parallel to NeRF and overfits a deep implicit network to a signal, allowing for arbitrary resolution features and low storage costs. In both architectures, our upsampled features can be drop-in replacements in downstream applications because our methods do not transform the semantics of the underlying features. We show that these upsampled features can significantly improve a variety of downstream tasks including semantic segmentation and depth prediction. Additionally, we show that model explanation methods such as CAM can be made higher-resolution using upsampled features. In particular, one can study a model's behavior with much greater detail without the need for complex methods based on relevance and information propagation \citep{relevancecam,infocam}. In summary, we include a short video describing FeatUp at \href{https://aka.ms/featup}{aka.ms/featup} and make the following contributions:

\begin{itemize}
    \item FeatUp: a new method to significantly improve the spatial resolution of any model's features, parametrized as either a fast feedforward upsampling network or an implicit network.
    \item A fast CUDA implementation of Joint Bilateral Upsampling orders of magnitude more efficient than a standard PyTorch implementation and allowing guided upsampling in large-scale models.
    \item We show that FeatUp features can be used as drop-in replacements for ordinary features to improve performance on dense prediction tasks and model explainability.
\end{itemize}

\section{Related Work}
\label{sec:related}
\vspace{-0.05in}

\begin{figure*}[t]
    \vspace{-.15in}
    \centering
    \includegraphics[width=0.95\linewidth]{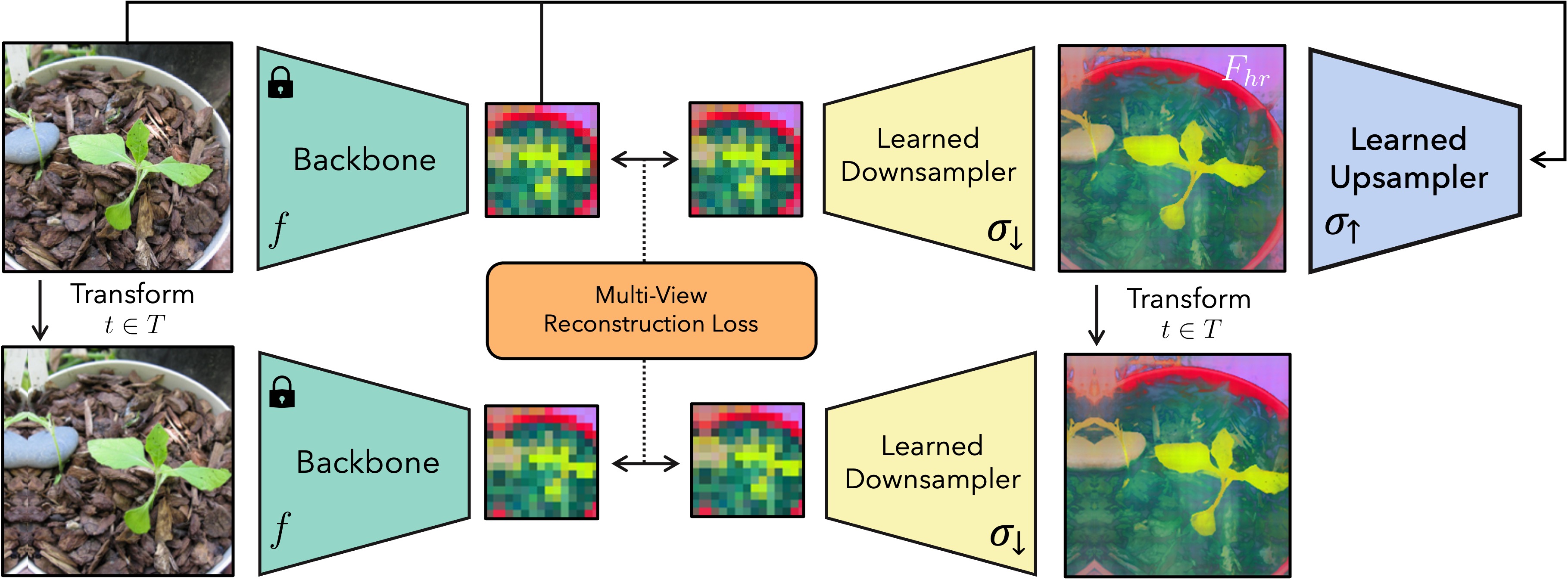}
    \caption{\small  The FeatUp training architecture. FeatUp learns to upsample features through a consistency loss on low resolution ``views'' of a model's features that arise from slight transformations of the input image.}
    \label{fig:jitter_diagram}
    \vspace{-.15in}
\end{figure*}

\begin{figure*}[t]
    \vspace{-.15in}
    \includegraphics[width=\linewidth]{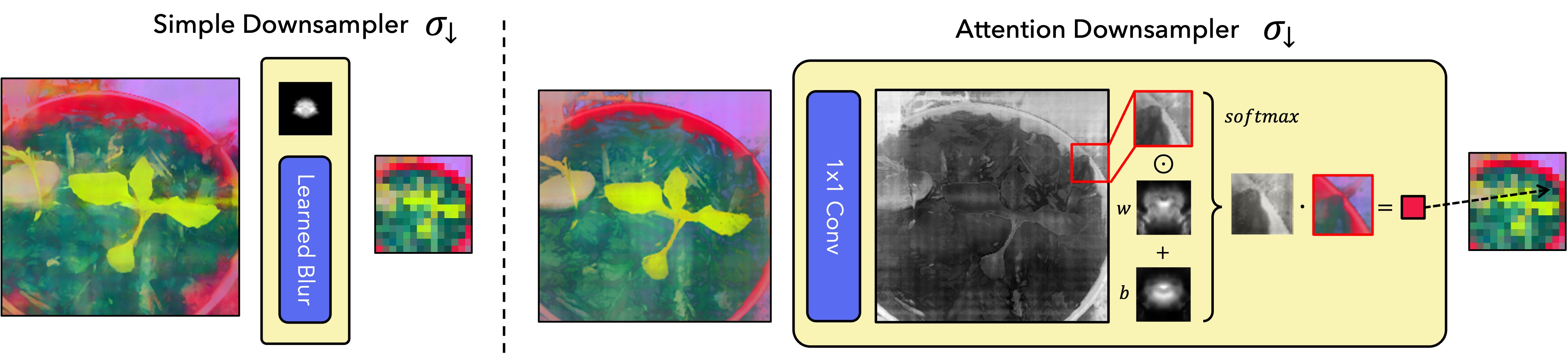}
    \vspace{-.25in}
    \caption{\small  We introduce two learned downsamplers. The simple downsampler (Left) is a fast learned blur kernel. The attention downsampler (right) combines a predicted salience map with spatially invariant kernels. This downsampler can better adapt to networks with nonlinear and dynamic receptive fields.
    }
    \label{fig:downsampler_comp}
    \vspace{-.1in}
\end{figure*}
\paragraph{Image-adaptive filtering.}
Adaptive filters are commonly used to enhance images while preserving their underlying structure and content.
For example, bilateral filters \citep{bilateral, guided_bilateral, joint_bilateral_dehaze} apply a spatial filter to a low-resolution signal and an intensity filter to a high-resolution guidance to blend information from the two. Joint Bilateral Upsampling (JBU) \citep{jbu} uses this technique to upsample a low-resolution signal with a high-resolution guidance. JBU has been used successfully for efficient image enhancement and other applications. Recently, some works embed bilateral filtering approaches \citep{gum} and nonlocal means \citep{nonlocal} into convolutional networks \citep{bilateral_inception, nonlocal_cnn, gharbi2017deep, Li:2018:DPI} and vision transformers \citep{arm, esrt}. Shape Recipes \citep{freeman2002shape} learn the local relationship between signals to create up-sample target signals. Pixel-adaptive convolutional (PAC) networks \citep{pac} adapt a convolution operation to input data and has been used to advance performance in segmentation \citep{single_stage_seg, cell_seg} and monocular depth estimation \citep{guided_monocular_depth, selfdeco, adaptive_depth}. The Spatially-Adaptive Convolution (SAC) in \citep{squeeze_seg} factorizes the adaptive filter into an attention map and convolution kernel. \citep{gadde2016superpixel} extend bilateral filtering to superpixels and embed this operation inside of a deep network to improve semantic segmentation. This class of methods, effective across a variety of applications, directly incorporates spatial information into the task while still allowing for flexibility in learning a network. 
\vspace{-0.1in}
\paragraph{Image super-resolution.}
One of the earliest deep unsupervised super-resolution methods was Zero-Shot Super-resolution (ZSSR) \citep{zssr}, which learns a single-image network at test time. Local implicit models \citep{LIIF} use locally-adaptive models to interpolate information, and have been shown to improve the performance of super-resolution networks. Deep Image Priors \citep{Ulyanov_2020} show that CNNs provide inductive biases for inverse problems such as zero-shot image denoising and super-resolution. While there is extensive literature on image super-resolution, these methods are not well-adapted to handle ultra-low resolution, yet high-dimensional deep features as we show in the Supplement. 
\vspace{-0.1in}
\paragraph{General-purpose feature upsampling.}
A widely-used approach to upsample deep feature maps is bilinear interpolation. Though efficient, this method blurs information and is insensitive to the content or the high-resolution structure in the original image. Nearest neighbor and bicubic interpolation \citep{keys1981cubic} have similar drawbacks. Evaluating a network on larger inputs can achieve higher resolutions but with a steep computational cost. Furthermore, this often degrades model performance and semantics due to the decreased relative receptive field size. For deep convolutional networks, one popular technique is to set final convolution strides to 1 \citep{long2015fully,infocam}. However, this approach yields blurry features, as the model's receptive field is still large. Recent works using visual transformers \citep{vit_feats, vit_splice} perform a similar modification on input patch strides and interpolate positional encodings. Though simple and reasonably effective, this approach incurs a steep increase in computational footprint for every $2\times$ increase in resolution, making it impossible to use in practice for larger upsampling factors. This approach can also distort features because of the previously mentioned fixed receptive field of the patches.
\vspace{-0.1in}
\paragraph{Image-adaptive feature upsampling.}
Many different operations exist in the literature to create features at higher resolutions. Deconvolutions \citep{conv_deconv, conv_guide, deconv, perceptual} and transposed convolutions \citep{conv_trans} use a learned kernel to transform features into a new space with a larger resolution. The resize-convolution \citep{checker_deconv} appends a learned convolution to deterministic upsampling procedure and reduces checkerboard artifacts that plague deconvolutions \citep{cgan_face, checker_deconv, superresolution}. The resize-convolution is now a common component of image decoders such as the U-Net \citep{unet} and has been applied to semantic segmentation \citep{hdenseunet, unet3p, contextual_deconv} and super-resolution \citep{lapsrn, sr_skip, srgan}. Other methods such as IndexNet \citep{indexnet} and Affinity-Aware Upsampling (A2U) \citep{dai2020learning} are effective on image matting but fall short on other dense prediction tasks \citep{lu2022fade}. Methods such as Pixel-Adaptive Convolutions \citep{pac}, CARAFE \citep{carafe}SAPA \cite{lu2022sapa}, and DGF \cite{wu2019fast} use learned input-adaptive operators to transform features. Though PAC is flexible, it does not upsample \textit{existing} feature maps faithfully and instead is used to transform features for downstream tasks. Additionally, DGF approximates the JBU operation with learned pointwise convolutions and linear maps, but does not fully implement JBU because the local query/model is computationally intractable. This is precisely the problem we solve exactly with our new efficient CUDA kernel.
Additionally, FADE \cite{lu2022fade} introduces a new semi-shift operator and uses decoder features to produce a joint feature upsampling module. \cite{ifa} view feature upsampling in a different light, focusing on a nearest-neighbors approach to align feature maps in encoder-decoder architectures with IFA. While IFA performs well on the specific semantic segmentation benchmarks, it does not take advantage of image guidance and fails to learn high quality representations outside of the encode-decoder framework, as we show in the Supplement.

\vspace{-0.07in}
\section{Methods}
\vspace{-0.1in}

\label{sec:methods}
The core intuition behind FeatUp is that one can compute high-resolution features by observing multiple different ``views'' of low-resolution features. We draw a comparison with 3D scene reconstruction models such as NeRF \citep{nerf}; in the same way that NeRF builds an implicit representation \citep{sitzmann2019siren,chen2019learning} of a 3D scene by enforcing consistency across many 2D photos of the scene, FeatUp builds an upsampler by enforcing consistency across many low-resolution feature maps. Like in broader NeRF literature, a variety of methods can arise from this basic idea. In this work, we introduce a lightweight, forward-pass upsampler based on Joint Bilateral Upsampling \citep{jbu} as well as an implicit network based upsampling strategy. The latter is learned per-image and query-able at arbitrary resolution. We provide an overview of the general FeatUp architecture in Figure \ref{fig:jitter_diagram}.

The first step in our pipeline is to generate low-resolution feature views to refine into a single high-resolution output. To this end, we perturb the input image with small pads, scales, and horizontal flips and apply the model to each transformed image to extract a collection of low-resolution feature maps. These small image jitters allow us to observe tiny differences in the output features and provide sub-feature information to train the upsampler.

Next, we construct a consistent high-resolution feature map from these views. We postulate that we can learn a latent high-resolution feature map that, when downsampled, reproduces our low-resolution jittered features (see Figure \ref{fig:jitter_diagram}). FeatUp's downsampling is a direct analog to ray-marching; just as 3D data is rendered into 2D in this NeRF step, our downsampler transforms high-resolution features into low-resolution features. Unlike NeRF, we do not need to estimate parameters that generate each view. Instead, we track the parameters used to ``jitter'' each image and apply \textit{the same} transformation to our learned high-resolution features prior to downsampling. We then compare downsampled features to the true model outputs using a gaussian likelihood loss \citep{hamilton2020likely}. A good high-resolution feature map should reconstruct the observed features across all the different views. 

\begin{figure*}[t]
    \includegraphics[width=\linewidth]{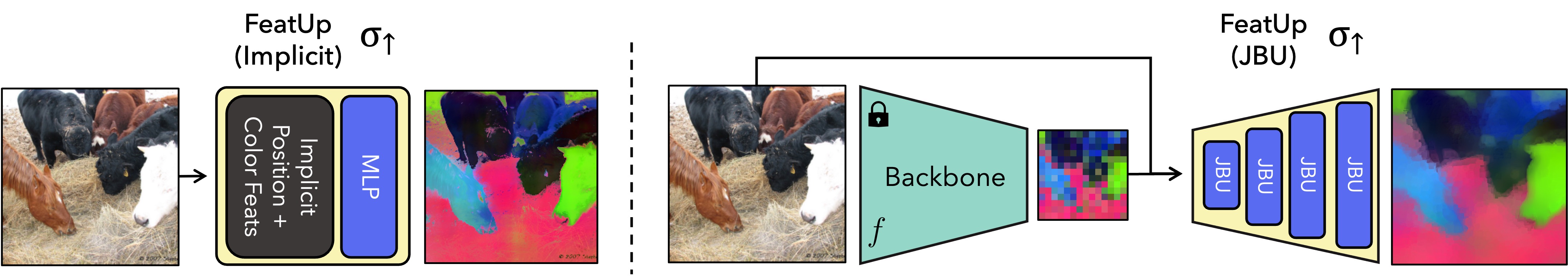}
    \vspace{-.25in}
    \caption{\small  
    Our Implicit version of FeatUp learns an implicit network to upsample a single image's features. Our JBU FeatUp learns a stack of JBUs that learns to quickly upsample features from a large image corpora. }
    \label{fig:upsampler_diagram}
   \vspace{-.15in}
\end{figure*}

More formally, let $t \in T$ be from a collection of small transforms such as pads, zooms, crops, horizontal flips, and their compositions. Let $x$ be an input image, $f$ be our model backbone, $\sigma_{\downarrow}$ be a learned downsampler, and $\sigma_{\uparrow}$ be a learned upsampler. We can form the predicted high-res features $F_{hr}$ by evaluating $F_{hr} = \sigma_{\uparrow}(f(x), x)$. We note that this parameterization allows $\sigma_{\uparrow}$ to be a guided upsampler (which depends on both $x$ and $f(x)$), an unguided upsampler (which depends on only $f(x)$), an implicit network (which depends on only $x$), or a learned buffer of features (which depends on nothing). We can now form our main multi-view reconstruction loss term as follows:

\vspace{-.15in}
\begin{equation}
    \mathcal{L}_{rec} = \frac{1}{|T|} \sum_{t \in T} \frac{1}{2s^2}\lVert f \left( t \left( x \right) \right) - \sigma_{\downarrow}\left(t \left( F_{hr} \right) \right) \rVert_2^2 + \log(s)
\end{equation}
\vspace{-.15in}

Where $\lVert \cdot \rVert$ is the standard squared $l_2$ norm and $s = \mathcal{N}(f \left( t \left( x \right) \right))$ is a spatially-varying adaptive uncertainty \citep{hamilton2020likely} parameterized by a small linear network $\mathcal{N}$. This turns the MSE loss into a proper likelihood capable of handling uncertainty. This extra flexibility allows the network to learn when certain outlier features fundamentally cannot be upsampled. In the supplement, we show this adaptive uncertainty's effectiveness in an ablation study and visualization. 

\begin{figure}[t]
    \centering
    \vspace{-.15in}
    \includegraphics[width=1.0\linewidth]{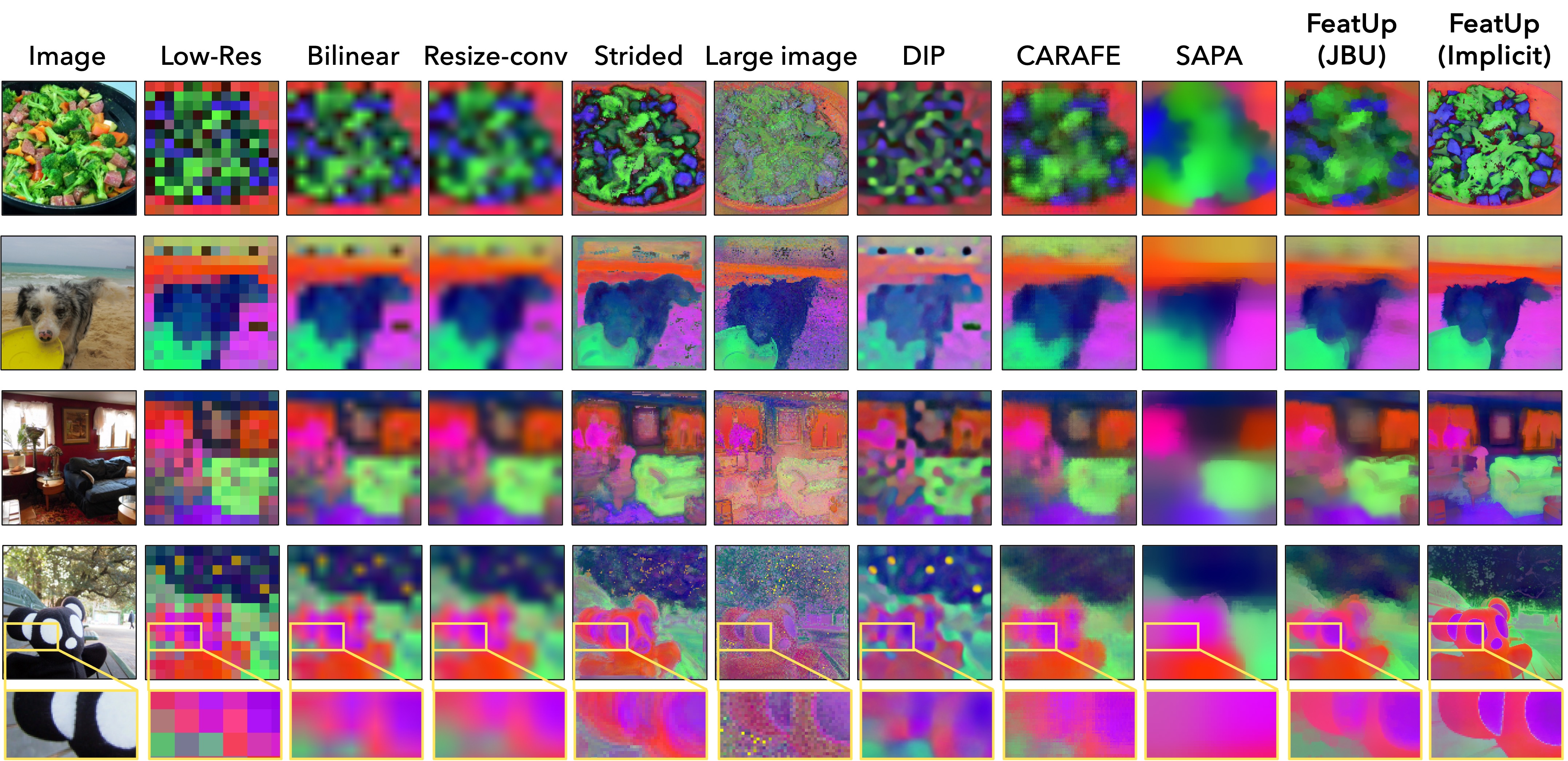}
    \vspace{-.15in}
    \caption{\small Low-res ViT features $(14\times14)$ from the COCO-Stuff validation set are upsampled by $16\times$. Bilinear and resize-conv baselines produce blurry outputs. Larger inputs and smaller transformer strides can help, but introduce noise or blur and are bound by time and memory constraints (We can only compute $8\times$ upsamplings for these methods, see Figure \ref{fig:memory_time}). Our FeatUp methods preserve semantics of the low-res features and recover lost spatial information from the high-res input image.
    }
   \vspace{-.15in}
    \label{fig:qualitative_upsampler_comp}
\end{figure}

\subsection{Choosing a Downsampler}
\label{sec:downsampler}
\vspace{-0.1in}

Our next architectural choice is the learned downsampler $\sigma_{\downarrow}$. We introduce two options: a fast and simple learned blur kernel, and a more flexible attention-based downsampler. Both proposed modules do not change the ``space'' or ``semantics'' of the features with nontrivial transformations, but rather only interpolate features within a small neighborhood. We diagram both choices in Figure \ref{fig:downsampler_comp} and demonstrate the effectiveness of the attention downsampler in Figure \ref{fig:ablation} of the Supplement.

Our simple downsampler blurs the features with a learned blur kernel and can be implemented as a convolution applied independently to each channel. The learned kernel is normalized to be non-negative and sum to 1 to ensure the features remain in the same space.

Though this blur-based downsampler is efficient, it cannot capture dynamic receptive fields, object salience, or other nonlinear effects. To this end, we also introduce a more flexible attention downsampler that spatially adapts the downsampling kernel. In short, this component uses a 1x1 convolution to predict a saliency map from the high-resolution features. It combines this saliency map with learned spatially-invariant weight and bias kernels and normalizes the result to create a spatially-varying blur kernel that interpolates the features. More formally:

\vspace{-.1in}
\begin{equation}
    \sigma_{\downarrow}(F_{hr})_{ij} = \text{softmax}( w \odot \text{Conv}(F_{hr}[\Omega_{ij}]) + b) \cdot F_{hr}[\Omega_{ij}] 
\end{equation}
\vspace{-.15in}

Where $\sigma_{\downarrow}(F)_{ij}$ is the $i,j$th component of the resulting feature map and $F_{hr}[\Omega_{ij}]$ refers to a patch of high resolution features corresponding to the $i,j$ location in the downsampled features. $\odot$ and $\cdot$ refer to the elementwise and inner products respectively, and $w$ and $b$ are learned weight and bias kernels shared across all patches. Our main hyperparameter for both downsamplers is the kernel size, which should be larger for models with larger receptive fields such as convolutional nets. We defer discussion of model-specific hyperparameters to the Supplement.

\begin{figure*}[t]
    \vspace{-.15in}
    \centering
    \includegraphics[width=\linewidth]{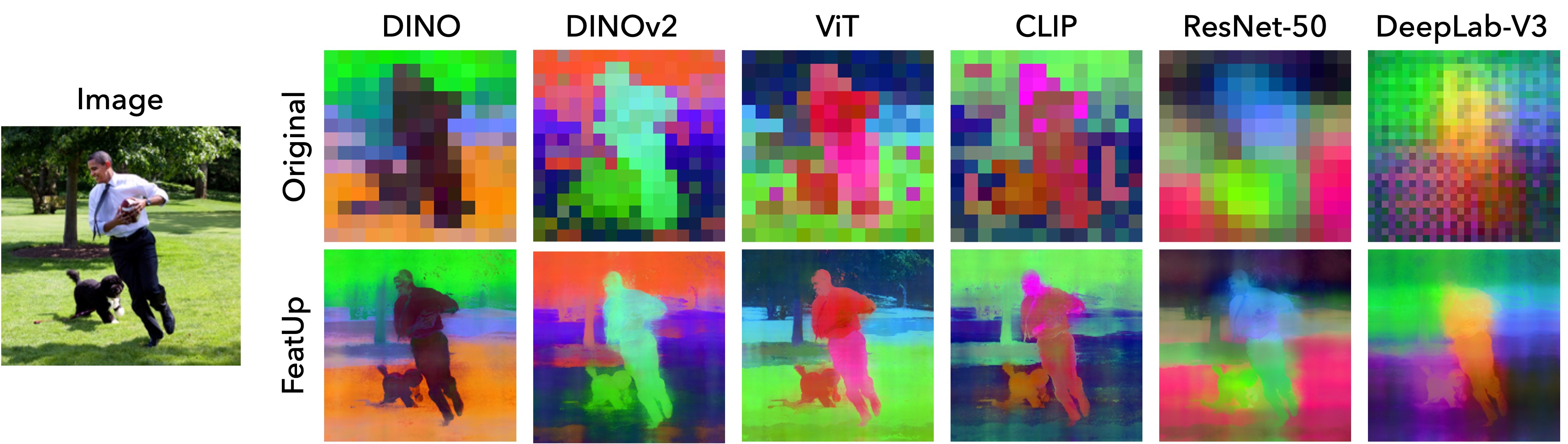}
    \vspace{-.15in}
    \caption{ \small  FeatUp can upsample the features of any backbone, even convnets with aggressive nonlinear pooling.}
    \label{fig:backbone_comp}
    \vspace{-.1in}

\end{figure*}

\subsection{Choosing an Upsampler}
\vspace{-0.1in}

A central choice in our architecture is the parameterization of $\sigma_{\uparrow}$. We introduce two variants: ``JBU'' FeatUp parameterizes $\sigma_{\uparrow}$ with a guided upsampler based on a stack of Joint Bilateral Upsamplers (JBU) \citep{jbu}. This architecture learns an upsampling strategy that generalizes across a corpus of images. The second method, ``Implicit'' FeatUp, uses an implicit network to parameterize $\sigma_{\uparrow}$ and can yield remarkably crisp features when overfit to a single image. Both methods are trained using the same broader architecture and loss. We illustrate both strategies in Figure \ref{fig:upsampler_diagram}.

 \paragraph{Joint Bilateral Upsampler.} Our feedforward upsampler uses a stack of parameterized joint bilateral upsamplers (JBU) \citep{jbu}:

\vspace{-.15in}
\begin{equation}
    F_{hr} = (\text{JBU}(\cdot, x) \circ  \text{JBU}(\cdot, x) \circ  ... )(f(x))
\end{equation}
\vspace{-.15in}

where $\circ$ is function composition, $f(x)$ is the low-resolution feature map, and $x$ is the original image. This architecture is fast, directly incorporates high-frequency details from the input image into the upsampling process, and is independent of the architecture of $f$. 
Our formulation generalizes the original JBU \citep{jbu} implementation to high-dimensional signals and makes this operation learnable. In joint bilateral upsampling we use a high-resolution signal, $G$, as guidance for the low-resolution features $F_{lr}$. We let $\Omega$ be a neighborhood of each pixel in the guidance. In practice, we use a $3\times3$ square centered at each pixel. Let $k(\cdot,\cdot)$ be a similarity kernel that measures how ``close'' two vectors are. We can then form our joint bilateral filter:

\vspace{-.15in}
\begin{equation}
    \label{eqn:ffu}
    \hat{F}_{hr}[i,j] = \frac{1}{Z} \sum_{(a,b) \in \Omega}  \biggl ( F_{lr}[a,b] \;\;
     k_{range}(G[i,j],G[a,b]) \;\; k_{spatial}([i,j], [a,b]) \biggr )
\end{equation}
\vspace{-.13in}

where $Z$ is a normalization factor to ensure the kernel sums to 1. Here, $k_{spatial}$ is a learnable Gaussian kernel on the Euclidean distance between coordinate vectors of width $\sigma_{spatial}$:

\vspace{-.12in}
\begin{equation}
         k_{spatial}(x,y) = \exp \left( \frac{-\lVert x - y\rVert_2^2}{2 \sigma_{spatial}^2} \right)
\end{equation}
\vspace{-.12in}

Furthermore, $k_{range}$ is a temperature-weighted softmax \citep{hamilton2020likely} applied to the inner products from a multi-layer perceptron (MLP) that operates on the guidance signal $G$:

\vspace{-.1in}
\begin{equation}
\label{eq:krange}
         k_{range}(x,y) = \text{softmax}_{(a,b) \in \Omega} \left( \frac{1}{\sigma_{range}^2} MLP(G[i,j]) \cdot MLP(G[a,b])  \right)
\end{equation}
\vspace{-.1in}

where $\sigma_{range}^2$ acts as the temperature. We note that the original JBU uses a fixed Gaussian kernel on the guidance signal, $G$. Our generalization performs much better as the MLP can be learned from data to create a better upsampler. In our experiments, we use a two-layer GeLU \citep{gelu} MLP with 30-dimensional hidden and output vectors. To evaluate $F_{lr}[a,b]$ we follow the original JBU formulation and use bilinear-interpolated features if the guidance pixel does not directly align with a low-resolution feature. For resolution independence, we use coordinate distances normalized to $[-1, 1]$ in the spatial kernel.

One challenge we faced was the poor speed and memory performance of existing JBU implementations. This could explain why this simple approach is not used more widely. To this end, we contribute an efficient CUDA implementation of the spatially adaptive kernel used in the JBU. Compared to a naive PyTorch implementation with the \texttt{torch.nn.Unfold} operator, our operation uses up to two orders of magnitude less memory and speeds inference by up to $10\times$. We demonstrate its significant performance improvements in Table \ref{tab:cuda_benchmark_full} of the supplement.

\paragraph{Implicit} 
Our second upsampler architecture draws a direct analogy with NeRF by parametrizing the high-resolution features of a single image with an implicit function $F_{hr} = \text{MLP}(z)$. Several existing upsampling solutions also take this inference-time training approach, including DIP \citep{Ulyanov_2020} and LIIF \citep{LIIF}. We use a small MLP to map image coordinates and intensities to a high-dimensional feature for the given location. We follow the guidance of prior works \citep{nerf, siren, fourier} and use Fourier features to improve the spatial resolution of our implicit representations. In addition to standard Fourier positional features, we show that adding Fourier color features allows the network to use high-frequency color information from the original image. This significantly speeds convergence and enables graceful use of high-resolution image information without techniques like Conditional Random Fields (CRFs). We illustrate the profound effect of Fourier color features in Section \ref{subsec:ablation} of the Supplement.

More formally, let $h(z, \hat{\omega})$ represent the component-wise discrete Fourier transform of an input signal $z$, with a vector of frequencies $\hat{\omega}$. Let $e_i$ and $e_j$ represent the two-dimensional pixel coordinate fields ranging in the interval $[-1, 1]$. Let $:$ represent concatenation along the channel dimension. We can now express our high-resolution feature map as:

\vspace{-.15in}
\begin{equation}
    F_{hr} = \text{MLP}(h(e_i:e_j:x, \hat{\omega}))
\end{equation}
\vspace{-.15in}

Our MLP is a small 3-layer ReLU \citep{glorot2011deep} network with dropout \citep{dropout}($p=.1$) and layer normalization \citep{ba2016layer}. We note that, at test time, we can query the pixel coordinate field to yield features $F_{hr}$ at \textbf{any} resolution. The number of parameters in our implicit representation is over two orders of magnitude smaller than a $(224 \times 224)$ explicit representation while being more expressive, significantly reducing convergence time and storage size.

\subsection{Additional Method Details}
\vspace{-0.05in}

\paragraph{Accelerated Training with Feature Compression} To reduce the memory footprint and further speed up the training of FeatUp's implicit network, we first compress the spatially-varying features to their top $k=128$ principal components. This operation is approximately lossless as the top 128 components explain $\sim 96\%$ of the variance across a single image's features. This improves training time by a factor of $60\times$ for ResNet-50, reduces the memory footprint, enables larger batches, and does not have any observable effect on learned feature quality. When training the JBU upsampler, we sample random projection matrices in each batch to avoid computing PCA in the inner loop. This achieves the same effect thanks to the Johnson–Lindenstrauss lemma \citep{johnson1986extensions}.

\paragraph{Total Variation Prior} To avoid spurious noise in the high resolution features, we add a small ($\lambda_{tv} = 0.05$) total variation smoothness prior \citep{rudin1992nonlinear} on the implicit feature magnitudes:

\vspace{-.15in}
\begin{equation}
    \mathcal{L}_{tv} = \sum_{i , j} \biggl ( (||F_{hr}[i,j]||-||F_{hr}[i-1,j]||)^2
    + (||F_{hr}[i,j]||-||F_{hr}[i,j-1]|| ) ^2 \biggr )
\end{equation}
\vspace{-.15in}

\begin{table}[t]
  \vspace{-.2in}
  \centering
  \renewcommand*{\arraystretch}{1.15}
\begin{tabular}{@{}rcccccc@{}}
\toprule
 & \multicolumn{2}{c}{CAM Score} & \multicolumn{2}{c}{Semantic Seg.} & \multicolumn{2}{c}{Depth Estimation} \\
 & A.D. $\downarrow$ & A.I. $\uparrow$ & Acc. $\uparrow$ & mIoU $\uparrow$ & RMSE $\downarrow$ & $\delta > 1.25$ $\uparrow$ \\ \midrule
Low-res & 10.69 & 4.81 & 65.17 & 40.65 & 1.25 & 0.894 \\
Bilinear & 10.24 & 4.91 & 66.95 & 42.40 & 1.19 & 0.910 \\
Resize-conv & 11.02 & 4.95 & 67.72 & 42.95 & 1.14 & 0.917 \\
DIP & 10.57 & 5.16 & 63.78 & 39.86 & 1.19 & 0.907 \\
Strided & 11.48 & 4.97 & 64.44 & 40.54 & 2.62 & 0.900 \\
Large image & 13.66 & 3.95 & 58.98 & 36.44 & 2.33 & 0.896 \\
CARAFE & 10.24 & 4.96 & 67.1 & 42.39 & \underline{1.09} & 0.920 \\
SAPA & 10.62 & 4.85 & 65.69 & 41.17 & 1.19 & 0.917 \\ \midrule
FeatUp (JBU) & \underline{9.83} & \underline{5.24} & \underline{68.77} & \underline{43.41} & \underline{1.09} & \textbf{0.938} \\
FeatUp (Implicit) & \textbf{8.84} & \textbf{5.60} & \textbf{71.58} & \textbf{47.37} & \textbf{1.04} & \underline{0.927} \\ \bottomrule
\end{tabular}
\vspace{-.05in}
\caption{\small Comparison of feature upsamplers across metrics on CAM faithfulness, linear probe semantic segmentation, and linear probe depth estimation. Both FeatUp variants consistently outperform other approaches, including other forward-pass upsamplers (CARAFE, SAPA) and features optimized at inference-time (DIP).}
  \label{tab:main_results}
  \vspace{-.05in}

\end{table}

\begin{figure*}[t]
    \vspace{-.15in}
    \centering
    \includegraphics[width=\linewidth]{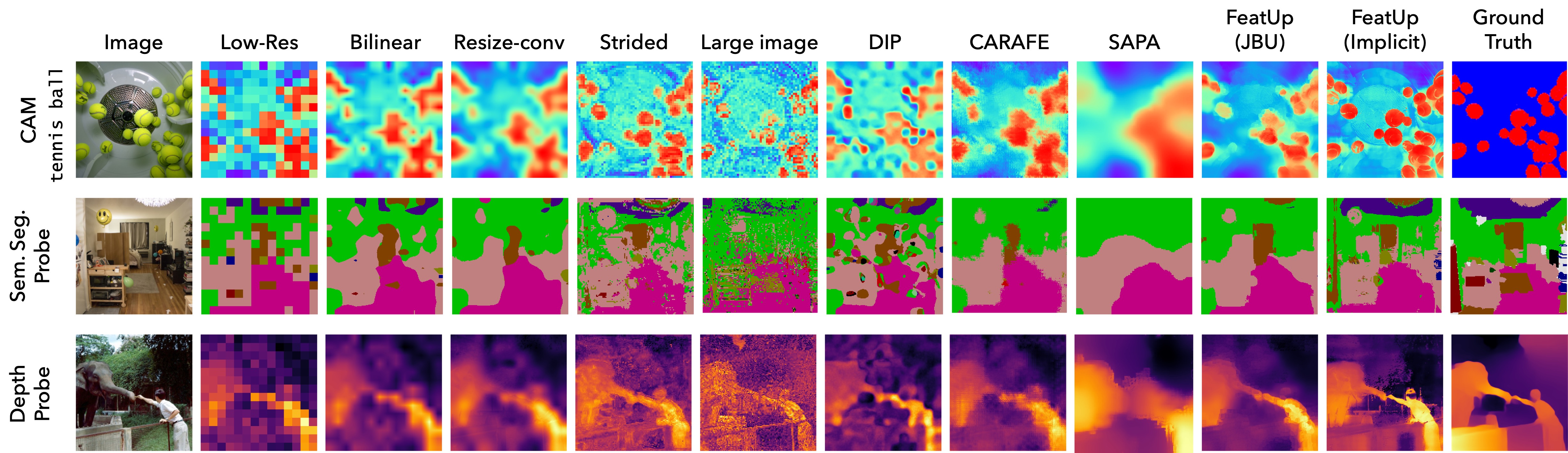}
    \vspace{-.2in}
    \caption{\small A comparison of different upsampling methods across each of the tasks considered in our analysis. FeatUp achieves significant improvements in resolution across each task.}
    \label{fig:main_results}
    \vspace{-.1in}
\end{figure*}

This is faster than regularizing full features and avoids overprescribing how the individual components should organize. We do not use this in the JBU upsampler because it does not suffer from overfitting. We demonstrate the importance of this regularizer in Section \ref{subsec:ablation} in the supplement.

\section{Experiments}
\label{sec:experiments}
\vspace{-0.15in}

We compare our method against several key upsampling baselines from the literature, in particular: Bilinear upsampling, Resize-conv, Strided (i.e. reducing the stride of the backbone's patch extractor), Large Image (i.e. using a larger input image), CARAFE \citep{carafe}, SAPA \citep{lu2022sapa}, and FADE \citep{lu2022fade}. We upsample ViT \citep{vit} features by $16\times$ (to the resolution of the input image) with every method except the strided and large-image baselines, which are computationally infeasible above $8\times$ upsampling. For additional details on the strided implementation, please refer to Section \ref{subsec:strided} of the Supplement.

\subsection{Qualitative Comparisons} 
\vspace{-0.05in}

\paragraph{Visualizing upsampling methods} Figure \ref{fig:qualitative_upsampler_comp} demonstrates the dramatic qualitative improvement FeatUp achieves compared to several baselines. Our visualizations fit a 3-dimensional PCA on each image's low-resolution ViT features and use this PCA to map upsampled features into the same RGB space. We also show that this high-fidelity upsampling extends to higher PCA components in Figure \ref{fig:pca_viz}, and that FeatUp can improve small object retrieval in Figure \ref{fig:supp_retrieval} in the Supplement.

\vspace{-0.05in}
\paragraph{Robustness across vision backbones} Figure \ref{fig:backbone_comp} demonstrates that FeatUp can upsample a variety of modern vision backbones. In particular, we show the implicit FeatUp features across a variety of backbones spanning transformers, convolutional nets, and both supervised and self-supervised models. Even though backbones like ResNet-50 do not precisely localize objects due to their large receptive fields, FeatUp can reasonably associate features to the correct object.

\subsection{Transfer Learning for Semantic Segmentation and Depth Estimation} 
\vspace{-0.1in}

Next, we demonstrate that FeatUp can serve as a drop-in replacement for existing features in downstream applications. To demonstrate this, we adopt the widely used experimental procedure of using linear probe transfer learning to evaluate representation quality. More specifically, we train linear probes on top of low-resolution features for both semantic segmentation and depth estimation. We then freeze and apply these probes to upsampled features to measure performance improvement. If features are valid drop-in improvements, existing probes should work well without adaptation. For all experiments, we use a frozen pre-trained ViT-S/16 as the featurizer, upsample the features (14x14 → 224x224), and extract maps by applying a linear layer on the features. 

\vspace{-0.05in}
For semantic segmentation, we follow the experimental setting of both \citep{probe,STEGO} and train a linear projection to predict the coarse classes of the COCO-Stuff (27 classes) training dataset using a cross-entropy loss. We report mIoU and accuracy on the validation set in Table \ref{tab:main_results}. For depth prediction we train on pseudo-labels from the MiDaS (DPT-Hybrid) \citep{ranftl2020towards} depth estimation network using their scale- and shift-invariant MSE. We report root mean square error (RMSE) and the $\delta > 1.25$ metric which is common in monocular depth estimation literature. More specifically this metric is defined as the percentage of pixels with $\delta = \max (\frac{y}{y^*}, \frac{y^*}{y}) > 1.25$ where $y$ is the depth prediction and $y^*$ is the ground truth.

We stress that these linear probe evaluations show that FeatUp features can improve downstream tasks \textit{without} re-training models. These analyses do not aim to create SOTA segmentation or depth networks. Both FeatUp variants outperform all baselines across all experiments, showing that either variant can be used as a drop-in replacement for existing features. Qualitatively, Figure \ref{fig:main_results} and Figures \ref{fig:supp_seg} - \ref{fig:supp_depth} in the supplement show cleaner, more cohesive predictions across both tasks.

\vspace{-0.05in}
\subsection{Class Activation Map Quality}
\vspace{-0.1in}

Attributing a model's predictions to specific pixels is crucial for diagnosing failures and understanding a model's behavior. Unfortunately, common interpretation methods like Class Activation Maps (CAM) are limited by the low res of the deep feature maps and cannot resolve small objects. We show that FeatUp features can be dropped into existing CAM analyses to yield stronger and more precise explanations. More specifically, we use the literature's established metrics, Average Drop (A.D.) and Average Increase (A.I.), that measure CAM quality (refer to Section \ref{subsec:adai} in the Supplement for a detailed description of these metrics). Intuitively, A.D. and A.I. capture how much an image's most salient region changes the classification output. A good CAM should highlight regions with the greatest effect on the classifier's predictions, so censoring these regions will have the largest impact on the model's predictions (lower A.D., higher A.I.). 
Upsamplers are trained on the ImageNet training set for 2,000 steps, and we compute metrics across 2,000 random images from the validation set. We use a frozen pre-trained ViT-S/16 as the featurizer, and extract CAMs by applying a linear classifier after max-pooling. Upsampling is done (14x14 → 224x224) on the features themselves, and CAMs are obtained from these high-resolution maps.
We report results in Table \ref{tab:main_results}, and Figures \ref{fig:main_results}, \ref{fig:supp_cam_vit}.

\vspace{-0.1in}
\subsection{End-to-end Semantic Segmentation}
\vspace{-0.1in}

FeatUp not only improves the resolution of pre-trained features but can also improve models learned end-to-end. We adopt the experimental setting of \citep{lu2022sapa,lu2022fade} to show that our JBU upsampler improves end-to-end performance on ADE20K semantic segmentation using the Segformer \citep{segformer} architecture. 
{Specifically, we train SegFormer on ADE20k \cite{ade1,ade2} (20,210 training and 2,000 val) for 160k steps. To validate that our setup matches that of existing literature despite numerical discrepancies, we also compute FLOPs for SegFormer with various upsamplers in Table \ref{tab:e2e_seg}. These counts are comparable with those in \cite{seg_flops}, confirming our architectural setup.
We report mean IoU, mean class accuracy (mAcc), and all-pixel accuracy (aAcc) against several recent baselines in Table \ref{tab:e2e_seg} including IndexNet \citep{indexnet}, A2U \citep{a2u}, CARAFE \citep{carafe}, SAPA \citep{lu2022sapa}, and FADE \citep{lu2022fade} in addition to more standard bilinear and resize-conv operators. Figure \ref{fig:seg_viz} in the Supplement shows examples of segmentation predictions across these methods. FeatUp consistently outperforms baselines with fewer added parameters, showing that FeatUp can also improve a broader, jointly trained architecture.

\begin{table}[t]
  \vspace{-.15in}
\centering
\setlength{\tabcolsep}{5pt}
  \renewcommand*{\arraystretch}{1.15}
\begin{tabular}{@{}ccccccccc@{}}
\toprule
 Metric & Bilinear & \begin{tabular}[c]{@{}c@{}}Resize-\\ conv\end{tabular} & IndexNet & A2U & CARAFE & SAPA & FADE & \begin{tabular}[c]{@{}c@{}}FeatUp\\ (JBU)\end{tabular} \\ \midrule
mIoU & 39.7 & 41.1 & 41.5 & 41.5 & 42.4 & 41.6 & \underline{43.6} & \textbf{44.2} \\
mAcc & 51.6 & 51.9 & 52.2 & 52.3 & 53.2 & \underline{55.3} & 54.8 & \textbf{55.8} \\
aAcc & 78.7 & 79.8 & \underline{80.2} & 79.9 & 80.1 & 79.8 & \textbf{80.7} & \textbf{80.7} \\ \midrule
\multicolumn{1}{l}{\begin{tabular}[c]{@{}r@{}} Params (M)\end{tabular}} & 13.7 & +3.54 & +12.6 & +0.12 & +0.78 & +0.20 & +0.29 & +0.16 \\
\multicolumn{1}{l}{\begin{tabular}[c]{@{}r@{}} GFLOPs \end{tabular}} & 16.0 & +34.40 & +30.90 & +0.51 & +1.66 & +1.15 & +2.95 & +1.70 \\ \bottomrule
\end{tabular}
\vspace{-.05in}

\caption{\small Semantic segmentation results with the Segformer \cite{segformer} architecture trained on the ADE20k train set and evaluated on the val set. FeatUp (JBU) outperforms the standard bilinear and resize-conv upsamplers in U-Net architectures, IndexNet \cite{indexnet}, A2U \cite{dai2020learning}, and other task-agnostic upsamplers (CARAFE \cite{carafe}, SAPA \cite{lu2022sapa}, FADE \cite{lu2022fade}). Additionally, our upsampler is competitive in parameter and floating-point operation count.}
\label{tab:e2e_seg}
\vspace{-.05in}

\end{table}



\vspace{-0.05in}
\section{Conclusion}
\label{sec:conc}
\vspace{-0.1in}
We present FeatUp, a novel approach to upsample deep features using multiview consistency. FeatUp solves a critical problem in computer vision: deep models learn high quality features but at prohibitively low spatial resolutions. Our JBU-based upsampler imposes strong spatial priors to accurately recover lost spatial information with a fast feedforward network based on a novel generalization of Joint Bilateral Upsampling. Our implicit FeatUp can learn high quality features at arbitrary resolutions. Both variants dramatically outperform a wide range of baselines across linear probe transfer learning, model interpretability, and end-to-end semantic segmentation.
\newpage
\section*{Acknowledgements}
\label{sec:conclusion}
\vspace{-0.1in}

We would like to thank the Microsoft Research Grand Central Resources team for their gracious help performing the experiments in this work. Special thanks to Oleg Losinets and Lifeng Li for their consistent, gracious, and timely help, debugging, and expertise. Without them, none of the experiments could have been run. 

This material is based upon work supported by the National Science Foundation Graduate Research Fellowship under Grant No. 2021323067. Any opinion, findings, and conclusions or recommendations expressed in this material are those of the authors(s) and do not necessarily reflect the views of the National Science Foundation. This research is based upon work supported in part by the Office of the Director of National Intelligence (Intelligence Advanced Research Projects Activity) via 2021-20111000006. The views and conclusions contained herein are those of the authors and should not be interpreted as necessarily representing the official policies, either expressed or implied, of ODNI, IARPA, or the U S Government. The US Government is authorized to reproduce and distribute reprints for governmental purposes notwithstanding any copyright annotation therein. This work is supported by the National Science Foundation under Cooperative Agreement PHY-2019786 (The NSF AI Institute for Artificial Intelligence and Fundamental Interactions, http://iaifi.org/)
Research was sponsored by the United States Air Force Research Laboratory and the United States Air Force Artificial Intelligence Accelerator and was accomplished under Cooperative Agreement Number FA8750-19-2- 1000. The views and conclusions contained in this document are those of the authors and should not be interpreted as representing the official policies, either expressed or implied, of the United States Air Force or the U.S. Government. The U.S. Government is authorized to reproduce and distribute reprints for Government purposes notwithstanding any copyright notation herein.

\bibliography{iclr2024_conference}

\begin{thebibliography}{91}
\providecommand{\natexlab}[1]{#1}
\providecommand{\url}[1]{\texttt{#1}}
\expandafter\ifx\csname urlstyle\endcsname\relax
  \providecommand{\doi}[1]{doi: #1}\else
  \providecommand{\doi}{doi: \begingroup \urlstyle{rm}\Url}\fi

\bibitem[Ahn et~al.(2019)Ahn, Cho, and Kwak]{ahn2019weakly}
Jiwoon Ahn, Sunghyun Cho, and Suha Kwak.
\newblock Weakly supervised learning of instance segmentation with inter-pixel relations.
\newblock In \emph{Proceedings of the IEEE/CVF conference on computer vision and pattern recognition}, pp.\  2209--2218, 2019.

\bibitem[Alain \& Bengio(2016)Alain and Bengio]{probe}
Guillaume Alain and Yoshua Bengio.
\newblock Understanding intermediate layers using linear classifier probes, 2016.
\newblock URL \url{https://arxiv.org/abs/1610.01644}.

\bibitem[Amir et~al.(2021)Amir, Gandelsman, Bagon, and Dekel]{vit_feats}
Shir Amir, Yossi Gandelsman, Shai Bagon, and Tali Dekel.
\newblock Deep vit features as dense visual descriptors, 2021.
\newblock URL \url{https://arxiv.org/abs/2112.05814}.

\bibitem[Araslanov \& Roth(2020)Araslanov and Roth]{single_stage_seg}
Nikita Araslanov and Stefan Roth.
\newblock Single-stage semantic segmentation from image labels.
\newblock In \emph{Proceedings of the IEEE/CVF Conference on Computer Vision and Pattern Recognition (CVPR)}, June 2020.

\bibitem[Ba et~al.(2016)Ba, Kiros, and Hinton]{ba2016layer}
Jimmy~Lei Ba, Jamie~Ryan Kiros, and Geoffrey~E Hinton.
\newblock Layer normalization.
\newblock \emph{arXiv preprint arXiv:1607.06450}, 2016.

\bibitem[Buades et~al.(2005)Buades, Coll, and Morel]{nonlocal}
A.~Buades, B.~Coll, and J.-M. Morel.
\newblock A non-local algorithm for image denoising.
\newblock In \emph{2005 IEEE Computer Society Conference on Computer Vision and Pattern Recognition (CVPR'05)}, volume~2, pp.\  60--65 vol. 2, 2005.
\newblock \doi{10.1109/CVPR.2005.38}.

\bibitem[Caraffa et~al.(2015)Caraffa, Tarel, and Charbonnier]{guided_bilateral}
Laurent Caraffa, Jean-Philippe Tarel, and Pierre Charbonnier.
\newblock The guided bilateral filter: When the joint/cross bilateral filter becomes robust.
\newblock \emph{IEEE Transactions on Image Processing}, 24\penalty0 (4):\penalty0 1199--1208, 2015.
\newblock \doi{10.1109/TIP.2015.2389617}.

\bibitem[Caron et~al.(2021)Caron, Touvron, Misra, J\'egou, Mairal, Bojanowski, and Joulin]{dino}
Mathilde Caron, Hugo Touvron, Ishan Misra, Herv\'e J\'egou, Julien Mairal, Piotr Bojanowski, and Armand Joulin.
\newblock Emerging properties in self-supervised vision transformers.
\newblock In \emph{Proceedings of the IEEE/CVF International Conference on Computer Vision (ICCV)}, pp.\  9650--9660, October 2021.

\bibitem[Chen et~al.(2021)Chen, Liu, and Wang]{LIIF}
Yinbo Chen, Sifei Liu, and Xiaolong Wang.
\newblock Learning continuous image representation with local implicit image function.
\newblock In \emph{Proceedings of the IEEE/CVF conference on computer vision and pattern recognition}, pp.\  8628--8638, 2021.

\bibitem[Chen \& Zhang(2019)Chen and Zhang]{chen2019learning}
Zhiqin Chen and Hao Zhang.
\newblock Learning implicit fields for generative shape modeling.
\newblock In \emph{Proceedings of the IEEE/CVF Conference on Computer Vision and Pattern Recognition}, pp.\  5939--5948, 2019.

\bibitem[Choi et~al.(2020)Choi, Lee, Kim, Kim, Kim, Sohn, and Min]{adaptive_depth}
Hyesong Choi, Hunsang Lee, Sunkyung Kim, Sunok Kim, Seungryong Kim, Kwanghoon Sohn, and Dongbo Min.
\newblock Adaptive confidence thresholding for monocular depth estimation, 2020.
\newblock URL \url{https://arxiv.org/abs/2009.12840}.

\bibitem[Choi et~al.(2021)Choi, Jung, Lee, Kim, Manocha, and Lee]{selfdeco}
Jaehoon Choi, Dongki Jung, Yonghan Lee, Deokhwa Kim, Dinesh Manocha, and Donghwan Lee.
\newblock Selfdeco: Self-supervised monocular depth completion in challenging indoor environments.
\newblock In \emph{2021 IEEE International Conference on Robotics and Automation (ICRA)}, pp.\  467--474, 2021.
\newblock \doi{10.1109/ICRA48506.2021.9560831}.

\bibitem[Dai et~al.(2020)Dai, Lu, and Shen]{dai2020learning}
Yutong Dai, Hao Lu, and Chunhua Shen.
\newblock Learning affinity-aware upsampling for deep image matting, 2020.

\bibitem[Dai et~al.(2021)Dai, Lu, and Shen]{a2u}
Yutong Dai, Hao Lu, and Chunhua Shen.
\newblock Learning affinity-aware upsampling for deep image matting.
\newblock In \emph{Proceedings of the IEEE/CVF Conference on Computer Vision and Pattern Recognition}, pp.\  6841--6850, 2021.

\bibitem[Dalal \& Triggs(2005)Dalal and Triggs]{hog}
N.~Dalal and B.~Triggs.
\newblock Histograms of oriented gradients for human detection.
\newblock In \emph{2005 IEEE Computer Society Conference on Computer Vision and Pattern Recognition (CVPR'05)}, volume~1, pp.\  886--893 vol. 1, 2005.
\newblock \doi{10.1109/CVPR.2005.177}.

\bibitem[Devlin et~al.(2018)Devlin, Chang, Lee, and Toutanova]{devlin2018bert}
Jacob Devlin, Ming-Wei Chang, Kenton Lee, and Kristina Toutanova.
\newblock Bert: Pre-training of deep bidirectional transformers for language understanding.
\newblock \emph{arXiv preprint arXiv:1810.04805}, 2018.

\bibitem[Dong et~al.(2015)Dong, Loy, He, and Tang]{superresolution}
Chao Dong, Chen~Change Loy, Kaiming He, and Xiaoou Tang.
\newblock Image super-resolution using deep convolutional networks, 2015.
\newblock URL \url{https://arxiv.org/abs/1501.00092}.

\bibitem[Dosovitskiy et~al.(2020)Dosovitskiy, Beyer, Kolesnikov, Weissenborn, Zhai, Unterthiner, Dehghani, Minderer, Heigold, Gelly, et~al.]{vit}
Alexey Dosovitskiy, Lucas Beyer, Alexander Kolesnikov, Dirk Weissenborn, Xiaohua Zhai, Thomas Unterthiner, Mostafa Dehghani, Matthias Minderer, Georg Heigold, Sylvain Gelly, et~al.
\newblock An image is worth 16x16 words: Transformers for image recognition at scale.
\newblock \emph{arXiv preprint arXiv:2010.11929}, 2020.

\bibitem[Dumoulin \& Visin(2016{\natexlab{a}})Dumoulin and Visin]{conv_guide}
Vincent Dumoulin and Francesco Visin.
\newblock A guide to convolution arithmetic for deep learning, 2016{\natexlab{a}}.
\newblock URL \url{https://arxiv.org/abs/1603.07285}.

\bibitem[Dumoulin \& Visin(2016{\natexlab{b}})Dumoulin and Visin]{conv_trans}
Vincent Dumoulin and Francesco Visin.
\newblock A guide to convolution arithmetic for deep learning.
\newblock \emph{arXiv preprint arXiv:1603.07285}, 2016{\natexlab{b}}.

\bibitem[Freeman \& Torralba(2002)Freeman and Torralba]{freeman2002shape}
William Freeman and Antonio Torralba.
\newblock Shape recipes: Scene representations that refer to the image.
\newblock \emph{Advances in Neural Information Processing Systems}, 15, 2002.

\bibitem[Fu et~al.(2020)Fu, Liu, Li, Bao, Yan, Fang, and Lu]{contextual_deconv}
Jun Fu, Jing Liu, Yong Li, Yongjun Bao, Weipeng Yan, Zhiwei Fang, and Hanqing Lu.
\newblock Contextual deconvolution network for semantic segmentation.
\newblock \emph{Pattern Recognition}, 101:\penalty0 107152, 2020.
\newblock ISSN 0031-3203.
\newblock \doi{https://doi.org/10.1016/j.patcog.2019.107152}.
\newblock URL \url{https://www.sciencedirect.com/science/article/pii/S0031320319304534}.

\bibitem[Gadde et~al.(2015)Gadde, Jampani, Kiefel, Kappler, and Gehler]{bilateral_inception}
Raghudeep Gadde, Varun Jampani, Martin Kiefel, Daniel Kappler, and Peter~V. Gehler.
\newblock Superpixel convolutional networks using bilateral inceptions, 2015.
\newblock URL \url{https://arxiv.org/abs/1511.06739}.

\bibitem[Gadde et~al.(2016)Gadde, Jampani, Kiefel, Kappler, and Gehler]{gadde2016superpixel}
Raghudeep Gadde, Varun Jampani, Martin Kiefel, Daniel Kappler, and Peter~V Gehler.
\newblock Superpixel convolutional networks using bilateral inceptions.
\newblock In \emph{Computer Vision--ECCV 2016: 14th European Conference, Amsterdam, The Netherlands, October 11--14, 2016, Proceedings, Part I 14}, pp.\  597--613. Springer, 2016.

\bibitem[Gauthier(2015)]{cgan_face}
Jon Gauthier.
\newblock Conditional generative adversarial nets for convolutional face generation.
\newblock 2015.

\bibitem[Gharbi et~al.(2017)Gharbi, Chen, Barron, Hasinoff, and Durand]{gharbi2017deep}
Micha{\"e}l Gharbi, Jiawen Chen, Jonathan~T Barron, Samuel~W Hasinoff, and Fr{\'e}do Durand.
\newblock Deep bilateral learning for real-time image enhancement.
\newblock \emph{ACM Transactions on Graphics (TOG)}, 36\penalty0 (4):\penalty0 118, 2017.

\bibitem[Glorot et~al.(2011)Glorot, Bordes, and Bengio]{glorot2011deep}
Xavier Glorot, Antoine Bordes, and Yoshua Bengio.
\newblock Deep sparse rectifier neural networks.
\newblock In \emph{Proceedings of the fourteenth international conference on artificial intelligence and statistics}, pp.\  315--323. JMLR Workshop and Conference Proceedings, 2011.

\bibitem[Guizilini et~al.(2020)Guizilini, Hou, Li, Ambrus, and Gaidon]{guided_monocular_depth}
Vitor Guizilini, Rui Hou, Jie Li, Rares Ambrus, and Adrien Gaidon.
\newblock Semantically-guided representation learning for self-supervised monocular depth, 2020.
\newblock URL \url{https://arxiv.org/abs/2002.12319}.

\bibitem[Hamilton et~al.(2020)Hamilton, Shelhamer, and Freeman]{hamilton2020likely}
Mark Hamilton, Evan Shelhamer, and William~T Freeman.
\newblock It is likely that your loss should be a likelihood.
\newblock \emph{arXiv preprint arXiv:2007.06059}, 2020.

\bibitem[Hamilton et~al.(2022)Hamilton, Zhang, Hariharan, Snavely, and Freeman]{STEGO}
Mark Hamilton, Zhoutong Zhang, Bharath Hariharan, Noah Snavely, and William~T Freeman.
\newblock Unsupervised semantic segmentation by distilling feature correspondences.
\newblock \emph{arXiv preprint arXiv:2203.08414}, 2022.

\bibitem[He et~al.(2015)He, Zhang, Ren, and Sun]{rn50}
Kaiming He, Xiangyu Zhang, Shaoqing Ren, and Jian Sun.
\newblock Deep residual learning for image recognition, 2015.
\newblock URL \url{https://arxiv.org/abs/1512.03385}.

\bibitem[He et~al.(2019)He, Fan, Wu, Xie, and Girshick]{moco}
Kaiming He, Haoqi Fan, Yuxin Wu, Saining Xie, and Ross Girshick.
\newblock Momentum contrast for unsupervised visual representation learning, 2019.
\newblock URL \url{https://arxiv.org/abs/1911.05722}.

\bibitem[Hendrycks \& Gimpel(2016)Hendrycks and Gimpel]{gelu}
Dan Hendrycks and Kevin Gimpel.
\newblock Gaussian error linear units (gelus).
\newblock \emph{arXiv preprint arXiv:1606.08415}, 2016.

\bibitem[Hsu et~al.(2021)Hsu, Bolte, Tsai, Lakhotia, Salakhutdinov, and Mohamed]{hubert}
Wei-Ning Hsu, Benjamin Bolte, Yao-Hung~Hubert Tsai, Kushal Lakhotia, Ruslan Salakhutdinov, and Abdelrahman Mohamed.
\newblock Hubert: Self-supervised speech representation learning by masked prediction of hidden units, 2021.
\newblock URL \url{https://arxiv.org/abs/2106.07447}.

\bibitem[Hu et~al.(2022)Hu, Chen, Xu, Borse, Cai, Porikli, and Wang]{ifa}
Hanzhe Hu, Yinbo Chen, Jiarui Xu, Shubhankar Borse, Hong Cai, Fatih Porikli, and Xiaolong Wang.
\newblock Learning implicit feature alignment function for semantic segmentation, 2022.

\bibitem[Huang et~al.(2020)Huang, Lin, Tong, Hu, Zhang, Iwamoto, Han, Chen, and Wu]{unet3p}
Huimin Huang, Lanfen Lin, Ruofeng Tong, Hongjie Hu, Qiaowei Zhang, Yutaro Iwamoto, Xianhua Han, Yen-Wei Chen, and Jian Wu.
\newblock Unet 3+: A full-scale connected unet for medical image segmentation.
\newblock In \emph{ICASSP 2020 - 2020 IEEE International Conference on Acoustics, Speech and Signal Processing (ICASSP)}, pp.\  1055--1059, 2020.
\newblock \doi{10.1109/ICASSP40776.2020.9053405}.

\bibitem[Johnson et~al.(2016)Johnson, Alahi, and Fei-Fei]{perceptual}
Justin Johnson, Alexandre Alahi, and Li~Fei-Fei.
\newblock Perceptual losses for real-time style transfer and super-resolution.
\newblock In Bastian Leibe, Jiri Matas, Nicu Sebe, and Max Welling (eds.), \emph{Computer Vision -- ECCV 2016}, pp.\  694--711, Cham, 2016. Springer International Publishing.

\bibitem[Johnson et~al.(1986)Johnson, Lindenstrauss, and Schechtman]{johnson1986extensions}
William~B Johnson, Joram Lindenstrauss, and Gideon Schechtman.
\newblock Extensions of lipschitz maps into banach spaces.
\newblock \emph{Israel Journal of Mathematics}, 54\penalty0 (2):\penalty0 129--138, 1986.

\bibitem[Keys(1981)]{keys1981cubic}
Robert Keys.
\newblock Cubic convolution interpolation for digital image processing.
\newblock \emph{IEEE transactions on acoustics, speech, and signal processing}, 29\penalty0 (6):\penalty0 1153--1160, 1981.

\bibitem[Kobayashi et~al.(2022)Kobayashi, Matsumoto, and Sitzmann]{kobayashi2022decomposing}
Sosuke Kobayashi, Eiichi Matsumoto, and Vincent Sitzmann.
\newblock Decomposing nerf for editing via feature field distillation.
\newblock \emph{arXiv preprint arXiv:2205.15585}, 2022.

\bibitem[Kopf et~al.(2007)Kopf, Cohen, Lischinski, and Uyttendaele]{jbu}
Johannes Kopf, Michael~F. Cohen, Dani Lischinski, and Matt Uyttendaele.
\newblock Joint bilateral upsampling.
\newblock \emph{ACM Trans. Graph.}, 26\penalty0 (3):\penalty0 96–es, jul 2007.
\newblock ISSN 0730-0301.
\newblock \doi{10.1145/1276377.1276497}.
\newblock URL \url{https://doi.org/10.1145/1276377.1276497}.

\bibitem[Lai et~al.(2017)Lai, Huang, Ahuja, and Yang]{lapsrn}
Wei-Sheng Lai, Jia-Bin Huang, Narendra Ahuja, and Ming-Hsuan Yang.
\newblock Deep laplacian pyramid networks for fast and accurate super-resolution, 2017.
\newblock URL \url{https://arxiv.org/abs/1704.03915}.

\bibitem[Ledig et~al.(2017)Ledig, Theis, Husz{\'a}r, Caballero, Cunningham, Acosta, Aitken, Tejani, Totz, Wang, et~al.]{srgan}
Christian Ledig, Lucas Theis, Ferenc Husz{\'a}r, Jose Caballero, Andrew Cunningham, Alejandro Acosta, Andrew Aitken, Alykhan Tejani, Johannes Totz, Zehan Wang, et~al.
\newblock Photo-realistic single image super-resolution using a generative adversarial network.
\newblock In \emph{Proceedings of the IEEE conference on computer vision and pattern recognition}, pp.\  4681--4690, 2017.

\bibitem[Lee et~al.(2021)Lee, Kim, Park, Eo, and Hwang]{relevancecam}
Jeong~Ryong Lee, Sewon Kim, Inyong Park, Taejoon Eo, and Dosik Hwang.
\newblock Relevance-cam: Your model already knows where to look.
\newblock In \emph{Proceedings of the IEEE/CVF Conference on Computer Vision and Pattern Recognition}, pp.\  14944--14953, 2021.

\bibitem[Li et~al.(2018{\natexlab{a}})Li, Gharbi, Adams, Durand, and Ragan-Kelley]{Li:2018:DPI}
Tzu-Mao Li, Micha{\"e}l Gharbi, Andrew Adams, Fr{\'e}do Durand, and Jonathan Ragan-Kelley.
\newblock Differentiable programming for image processing and deep learning in {Halide}.
\newblock \emph{ACM Trans. Graph. (Proc. SIGGRAPH)}, 37\penalty0 (4):\penalty0 139:1--139:13, 2018{\natexlab{a}}.

\bibitem[Li et~al.(2018{\natexlab{b}})Li, Chen, Qi, Dou, Fu, and Heng]{hdenseunet}
Xiaomeng Li, Hao Chen, Xiaojuan Qi, Qi~Dou, Chi-Wing Fu, and Pheng-Ann Heng.
\newblock H-denseunet: Hybrid densely connected unet for liver and tumor segmentation from ct volumes.
\newblock \emph{IEEE Transactions on Medical Imaging}, 37\penalty0 (12):\penalty0 2663--2674, 2018{\natexlab{b}}.
\newblock \doi{10.1109/TMI.2018.2845918}.

\bibitem[Liu et~al.(2010)Liu, Yuen, and Torralba]{liu2010sift}
Ce~Liu, Jenny Yuen, and Antonio Torralba.
\newblock Sift flow: Dense correspondence across scenes and its applications.
\newblock \emph{IEEE transactions on pattern analysis and machine intelligence}, 33\penalty0 (5):\penalty0 978--994, 2010.

\bibitem[Liu et~al.(2023)Liu, Lu, Fu, and Cao]{seg_flops}
Wenze Liu, Hao Lu, Hongtao Fu, and Zhiguo Cao.
\newblock Learning to upsample by learning to sample, 2023.

\bibitem[Long et~al.(2015)Long, Shelhamer, and Darrell]{long2015fully}
Jonathan Long, Evan Shelhamer, and Trevor Darrell.
\newblock Fully convolutional networks for semantic segmentation.
\newblock In \emph{Proceedings of the IEEE conference on computer vision and pattern recognition}, pp.\  3431--3440, 2015.

\bibitem[LoweDavid(2004)]{sift}
G~LoweDavid.
\newblock Distinctive image features from scale-invariant keypoints.
\newblock \emph{International Journal of Computer Vision}, 2004.

\bibitem[Lu et~al.(2022{\natexlab{a}})Lu, Dai, Shen, and Xu]{indexnet}
Hao Lu, Yutong Dai, Chunhua Shen, and Songcen Xu.
\newblock Index networks.
\newblock \emph{IEEE Transactions on Pattern Analysis and Machine Intelligence}, 44\penalty0 (1):\penalty0 242--255, 2022{\natexlab{a}}.
\newblock \doi{10.1109/TPAMI.2020.3004474}.

\bibitem[Lu et~al.(2022{\natexlab{b}})Lu, Liu, Fu, and Cao]{lu2022fade}
Hao Lu, Wenze Liu, Hongtao Fu, and Zhiguo Cao.
\newblock Fade: Fusing the assets of decoder and encoder for task-agnostic upsampling.
\newblock In \emph{Proc. European Conference on Computer Vision (ECCV)}, 2022{\natexlab{b}}.

\bibitem[Lu et~al.(2022{\natexlab{c}})Lu, Liu, Ye, Fu, Liu, and Cao]{lu2022sapa}
Hao Lu, Wenze Liu, Zixuan Ye, Hongtao Fu, Yuliang Liu, and Zhiguo Cao.
\newblock Sapa: Similarity-aware point affiliation for feature upsampling.
\newblock In \emph{Proc. Annual Conference on Neural Information Processing Systems (NeurIPS)}, 2022{\natexlab{c}}.

\bibitem[Lu et~al.(2022{\natexlab{d}})Lu, Li, Liu, Huang, Zhang, and Zeng]{esrt}
Zhisheng Lu, Juncheng Li, Hong Liu, Chaoyan Huang, Linlin Zhang, and Tieyong Zeng.
\newblock Transformer for single image super-resolution.
\newblock In \emph{Proceedings of the IEEE/CVF Conference on Computer Vision and Pattern Recognition (CVPR) Workshops}, pp.\  457--466, June 2022{\natexlab{d}}.

\bibitem[Mazzini(2018)]{gum}
Davide Mazzini.
\newblock Guided upsampling network for real-time semantic segmentation, 2018.
\newblock URL \url{https://arxiv.org/abs/1807.07466}.

\bibitem[Mikolov et~al.(2013)Mikolov, Chen, Corrado, and Dean]{mikolov2013efficient}
Tomas Mikolov, Kai Chen, Greg Corrado, and Jeffrey Dean.
\newblock Efficient estimation of word representations in vector space.
\newblock \emph{arXiv preprint arXiv:1301.3781}, 2013.

\bibitem[Mildenhall et~al.(2020)Mildenhall, Srinivasan, Tancik, Barron, Ramamoorthi, and Ng]{nerf}
Ben Mildenhall, Pratul~P. Srinivasan, Matthew Tancik, Jonathan~T. Barron, Ravi Ramamoorthi, and Ren Ng.
\newblock Nerf: Representing scenes as neural radiance fields for view synthesis, 2020.
\newblock URL \url{https://arxiv.org/abs/2003.08934}.

\bibitem[Noh et~al.(2015)Noh, Hong, and Han]{deconv}
Hyeonwoo Noh, Seunghoon Hong, and Bohyung Han.
\newblock Learning deconvolution network for semantic segmentation.
\newblock \emph{2015 IEEE International Conference on Computer Vision (ICCV)}, pp.\  1520--1528, 2015.

\bibitem[Odena et~al.(2016)Odena, Dumoulin, and Olah]{checker_deconv}
Augustus Odena, Vincent Dumoulin, and Chris Olah.
\newblock Deconvolution and checkerboard artifacts.
\newblock \emph{Distill}, 2016.
\newblock \doi{10.23915/distill.00003}.
\newblock URL \url{http://distill.pub/2016/deconv-checkerboard}.

\bibitem[Prangemeier et~al.(2020)Prangemeier, Reich, and Koeppl]{cell_seg}
Tim Prangemeier, Christoph Reich, and Heinz Koeppl.
\newblock Attention-based transformers for instance segmentation of cells in microstructures.
\newblock In \emph{2020 IEEE International Conference on Bioinformatics and Biomedicine (BIBM)}, pp.\  700--707, 2020.
\newblock \doi{10.1109/BIBM49941.2020.9313305}.

\bibitem[Qian et~al.(2021)Qian, Shao, Zhu, Li, and Jia]{arm}
Shengju Qian, Hao Shao, Yi~Zhu, Mu~Li, and Jiaya Jia.
\newblock Blending anti-aliasing into vision transformer.
\newblock In M.~Ranzato, A.~Beygelzimer, Y.~Dauphin, P.S. Liang, and J.~Wortman Vaughan (eds.), \emph{Advances in Neural Information Processing Systems}, volume~34, pp.\  5416--5429. Curran Associates, Inc., 2021.
\newblock URL \url{https://proceedings.neurips.cc/paper/2021/file/2b3bf3eee2475e03885a110e9acaab61-Paper.pdf}.

\bibitem[Qin et~al.(2019)Qin, Kim, and Gedeon]{infocam}
Zhenyue Qin, Dongwoo Kim, and Tom Gedeon.
\newblock Rethinking softmax with cross-entropy: Neural network classifier as mutual information estimator.
\newblock \emph{arXiv preprint arXiv:1911.10688}, 2019.

\bibitem[Radford \& Narasimhan(2018)Radford and Narasimhan]{gpt}
Alec Radford and Karthik Narasimhan.
\newblock Improving language understanding by generative pre-training.
\newblock 2018.

\bibitem[Ranftl et~al.(2020)Ranftl, Lasinger, Hafner, Schindler, and Koltun]{ranftl2020towards}
Ren{\'e} Ranftl, Katrin Lasinger, David Hafner, Konrad Schindler, and Vladlen Koltun.
\newblock Towards robust monocular depth estimation: Mixing datasets for zero-shot cross-dataset transfer.
\newblock \emph{IEEE transactions on pattern analysis and machine intelligence}, 2020.

\bibitem[Rombach et~al.(2021)Rombach, Blattmann, Lorenz, Esser, and Ommer]{StableDiffusion}
Robin Rombach, Andreas Blattmann, Dominik Lorenz, Patrick Esser, and Björn Ommer.
\newblock High-resolution image synthesis with latent diffusion models, 2021.
\newblock URL \url{https://arxiv.org/abs/2112.10752}.

\bibitem[Ronneberger et~al.(2015)Ronneberger, Fischer, and Brox]{unet}
Olaf Ronneberger, Philipp Fischer, and Thomas Brox.
\newblock U-net: Convolutional networks for biomedical image segmentation.
\newblock \emph{CoRR}, abs/1505.04597, 2015.
\newblock URL \url{http://arxiv.org/abs/1505.04597}.

\bibitem[Rudin et~al.(1992)Rudin, Osher, and Fatemi]{rudin1992nonlinear}
Leonid~I Rudin, Stanley Osher, and Emad Fatemi.
\newblock Nonlinear total variation based noise removal algorithms.
\newblock \emph{Physica D: nonlinear phenomena}, 60\penalty0 (1-4):\penalty0 259--268, 1992.

\bibitem[Schneider et~al.(2019)Schneider, Baevski, Collobert, and Auli]{wav2vec}
Steffen Schneider, Alexei Baevski, Ronan Collobert, and Michael Auli.
\newblock wav2vec: Unsupervised pre-training for speech recognition, 2019.
\newblock URL \url{https://arxiv.org/abs/1904.05862}.

\bibitem[Shao et~al.(2014)Shao, Zhu, and Li]{shao2014transfer}
Ling Shao, Fan Zhu, and Xuelong Li.
\newblock Transfer learning for visual categorization: A survey.
\newblock \emph{IEEE transactions on neural networks and learning systems}, 26\penalty0 (5):\penalty0 1019--1034, 2014.

\bibitem[Shi et~al.(2016)Shi, Caballero, Theis, Huszar, Aitken, Ledig, and Wang]{conv_deconv}
Wenzhe Shi, Jose Caballero, Lucas Theis, Ferenc Huszar, Andrew Aitken, Christian Ledig, and Zehan Wang.
\newblock Is the deconvolution layer the same as a convolutional layer?, 2016.
\newblock URL \url{https://arxiv.org/abs/1609.07009}.

\bibitem[Shocher et~al.(2018)Shocher, Cohen, and Irani]{zssr}
Assaf Shocher, Nadav Cohen, and Michal Irani.
\newblock Zero-shot super-resolution using deep internal learning.
\newblock In \emph{2018 IEEE/CVF Conference on Computer Vision and Pattern Recognition}, pp.\  3118--3126, 2018.
\newblock \doi{10.1109/CVPR.2018.00329}.

\bibitem[Sitzmann et~al.(2020{\natexlab{a}})Sitzmann, Martel, Bergman, Lindell, and Wetzstein]{siren}
Vincent Sitzmann, Julien N.~P. Martel, Alexander~W. Bergman, David~B. Lindell, and Gordon Wetzstein.
\newblock Implicit neural representations with periodic activation functions, 2020{\natexlab{a}}.
\newblock URL \url{https://arxiv.org/abs/2006.09661}.

\bibitem[Sitzmann et~al.(2020{\natexlab{b}})Sitzmann, Martel, Bergman, Lindell, and Wetzstein]{sitzmann2019siren}
Vincent Sitzmann, Julien~N.P. Martel, Alexander~W. Bergman, David~B. Lindell, and Gordon Wetzstein.
\newblock Implicit neural representations with periodic activation functions.
\newblock In \emph{Proc. NeurIPS}, 2020{\natexlab{b}}.

\bibitem[Srivastava et~al.(2014)Srivastava, Hinton, Krizhevsky, Sutskever, and Salakhutdinov]{dropout}
Nitish Srivastava, Geoffrey Hinton, Alex Krizhevsky, Ilya Sutskever, and Ruslan Salakhutdinov.
\newblock Dropout: a simple way to prevent neural networks from overfitting.
\newblock \emph{The journal of machine learning research}, 15\penalty0 (1):\penalty0 1929--1958, 2014.

\bibitem[Su et~al.(2019)Su, Jampani, Sun, Gallo, Learned{-}Miller, and Kautz]{pac}
Hang Su, Varun Jampani, Deqing Sun, Orazio Gallo, Erik~G. Learned{-}Miller, and Jan Kautz.
\newblock Pixel-adaptive convolutional neural networks.
\newblock \emph{CoRR}, abs/1904.05373, 2019.
\newblock URL \url{http://arxiv.org/abs/1904.05373}.

\bibitem[Tancik et~al.(2020)Tancik, Srinivasan, Mildenhall, Fridovich-Keil, Raghavan, Singhal, Ramamoorthi, Barron, and Ng]{fourier}
Matthew Tancik, Pratul~P. Srinivasan, Ben Mildenhall, Sara Fridovich-Keil, Nithin Raghavan, Utkarsh Singhal, Ravi Ramamoorthi, Jonathan~T. Barron, and Ren Ng.
\newblock Fourier features let networks learn high frequency functions in low dimensional domains, 2020.
\newblock URL \url{https://arxiv.org/abs/2006.10739}.

\bibitem[Teed \& Deng(2020)Teed and Deng]{teed2020raft}
Zachary Teed and Jia Deng.
\newblock Raft: Recurrent all-pairs field transforms for optical flow.
\newblock In \emph{European conference on computer vision}, pp.\  402--419. Springer, 2020.

\bibitem[Tomasi \& Manduchi(1998)Tomasi and Manduchi]{bilateral}
C.~Tomasi and R.~Manduchi.
\newblock Bilateral filtering for gray and color images.
\newblock In \emph{Sixth International Conference on Computer Vision (IEEE Cat. No.98CH36271)}, pp.\  839--846, 1998.
\newblock \doi{10.1109/ICCV.1998.710815}.

\bibitem[Tong et~al.(2017)Tong, Li, Liu, and Gao]{sr_skip}
Tong Tong, Gen Li, Xiejie Liu, and Qinquan Gao.
\newblock Image super-resolution using dense skip connections.
\newblock In \emph{Proceedings of the IEEE international conference on computer vision}, pp.\  4799--4807, 2017.

\bibitem[Tumanyan et~al.(2022)Tumanyan, Bar-Tal, Bagon, and Dekel]{vit_splice}
Narek Tumanyan, Omer Bar-Tal, Shai Bagon, and Tali Dekel.
\newblock Splicing vit features for semantic appearance transfer, 2022.

\bibitem[Ulyanov et~al.(2020)Ulyanov, Vedaldi, and Lempitsky]{Ulyanov_2020}
Dmitry Ulyanov, Andrea Vedaldi, and Victor Lempitsky.
\newblock Deep image prior.
\newblock \emph{International Journal of Computer Vision}, 128\penalty0 (7):\penalty0 1867--1888, mar 2020.
\newblock \doi{10.1007/s11263-020-01303-4}.
\newblock URL \url{https://doi.org/10.1007%2Fs11263-020-01303-4}.

\bibitem[Wang et~al.(2019)Wang, Chen, Xu, Liu, Loy, and Lin]{carafe}
Jiaqi Wang, Kai Chen, Rui Xu, Ziwei Liu, Chen~Change Loy, and Dahua Lin.
\newblock Carafe: Content-aware reassembly of features.
\newblock 2019.
\newblock \doi{10.48550/ARXIV.1905.02188}.
\newblock URL \url{https://arxiv.org/abs/1905.02188}.

\bibitem[Wang et~al.(2017)Wang, Girshick, Gupta, and He]{nonlocal_cnn}
Xiaolong Wang, Ross Girshick, Abhinav Gupta, and Kaiming He.
\newblock Non-local neural networks, 2017.
\newblock URL \url{https://arxiv.org/abs/1711.07971}.

\bibitem[Wang et~al.(2020)Wang, Zhang, Kan, Shan, and Chen]{wang2020self}
Yude Wang, Jie Zhang, Meina Kan, Shiguang Shan, and Xilin Chen.
\newblock Self-supervised equivariant attention mechanism for weakly supervised semantic segmentation.
\newblock In \emph{Proceedings of the IEEE/CVF Conference on Computer Vision and Pattern Recognition}, pp.\  12275--12284, 2020.

\bibitem[Weiss et~al.(2016)Weiss, Khoshgoftaar, and Wang]{weiss2016survey}
Karl Weiss, Taghi~M Khoshgoftaar, and DingDing Wang.
\newblock A survey of transfer learning.
\newblock \emph{Journal of Big data}, 3\penalty0 (1):\penalty0 1--40, 2016.

\bibitem[Wu et~al.(2019)Wu, Zheng, Zhang, and Huang]{wu2019fast}
Huikai Wu, Shuai Zheng, Junge Zhang, and Kaiqi Huang.
\newblock Fast end-to-end trainable guided filter, 2019.

\bibitem[Xiao \& Gan(2012)Xiao and Gan]{joint_bilateral_dehaze}
Chunxia Xiao and Jiajia Gan.
\newblock Fast image dehazing using guided joint bilateral filter.
\newblock \emph{Vis. Comput.}, 28\penalty0 (6–8):\penalty0 713–721, jun 2012.
\newblock ISSN 0178-2789.
\newblock \doi{10.1007/s00371-012-0679-y}.
\newblock URL \url{https://doi.org/10.1007/s00371-012-0679-y}.

\bibitem[Xie et~al.(2021)Xie, Wang, Yu, Anandkumar, Alvarez, and Luo]{segformer}
Enze Xie, Wenhai Wang, Zhiding Yu, Anima Anandkumar, Jose~M Alvarez, and Ping Luo.
\newblock Segformer: Simple and efficient design for semantic segmentation with transformers.
\newblock \emph{Advances in Neural Information Processing Systems}, 34:\penalty0 12077--12090, 2021.

\bibitem[Xu et~al.(2020)Xu, Wu, Wang, Zhan, Vajda, Keutzer, and Tomizuka]{squeeze_seg}
Chenfeng Xu, Bichen Wu, Zining Wang, Wei Zhan, Peter Vajda, Kurt Keutzer, and Masayoshi Tomizuka.
\newblock Squeezesegv3: Spatially-adaptive convolution for efficient point-cloud segmentation.
\newblock In Andrea Vedaldi, Horst Bischof, Thomas Brox, and Jan-Michael Frahm (eds.), \emph{Computer Vision -- ECCV 2020}, pp.\  1--19, Cham, 2020. Springer International Publishing.
\newblock ISBN 978-3-030-58604-1.

\bibitem[Zhou et~al.(2017)Zhou, Zhao, Puig, Fidler, Barriuso, and Torralba]{ade2}
Bolei Zhou, Hang Zhao, Xavier Puig, Sanja Fidler, Adela Barriuso, and Antonio Torralba.
\newblock Scene parsing through ade20k dataset.
\newblock In \emph{Proceedings of the IEEE Conference on Computer Vision and Pattern Recognition}, 2017.

\bibitem[Zhou et~al.(2019)Zhou, Zhao, Puig, Xiao, Fidler, Barriuso, and Torralba]{ade1}
Bolei Zhou, Hang Zhao, Xavier Puig, Tete Xiao, Sanja Fidler, Adela Barriuso, and Antonio Torralba.
\newblock Semantic understanding of scenes through the ade20k dataset.
\newblock \emph{International Journal of Computer Vision}, 127\penalty0 (3):\penalty0 302--321, 2019.

\end{thebibliography}
\bibliographystyle{iclr2024_conference}

\newpage
\section{Supplemental Information}
\label{sec:supplement}
\renewcommand{\arraystretch}{1.5}

\subsection{Website, Video, and Code}

We provide additional details and a short video explaining FeatUp at \href{https://aka.ms/featup}{aka.ms/featup}. Additionally, we provide our code at: \href{https://tinyurl.com/28h3yppa}{https://tinyurl.com/28h3yppa}

\subsection{Strided baseline implementation}
\label{subsec:strided}
For the DINO and ViT backbones, we extract patches with a stride of $\frac{16}{\text{upsample factor}}$ to produce a higher density of feature vectors and thus increase feature resolution. We point out that the upsampling factor is limited with this method (as the stride is lower bounded by 1), so this approach can only upsample up to 16x for ViT-S/16. Practically however, these maximum upsampling factors are impractical as they require far more memory than current GPUs provide (see Figure \ref{fig:memory_time}).

\subsection{Comparison to Image-Upsampling Methods}
A variety of methods have been proposed for image super-resolution. Among the learning-based approaches, deep image prior (DIP) \citep{Ulyanov_2020} has been used succesfully for enhancing images without additional training data. Figure \ref{fig:baselines} shows that DIP poorly upsamples features, introducing artifacts and ``blob" patterns in the features and downstream outputs. \citep{zssr} introduced Zero-Shot Super-Resolution, a method that learns an image-specific CNN at test time without additional training data. Additionally, images can be represented as Local Implicit Image Functions (LIIF) \citep{LIIF} which can be queried at arbitrary resolution. While similar to FeatUp's implicit network, LIIF trained to continuously represent a feature map does not produce sharp outputs like FeatUp (Figure \ref{fig:baselines}) Despite these methods' successes in the image super-resolution problem space, they are not equipped to upsample high-dimensional features.

\begin{figure}[H]
\centering
    \includegraphics[width=\linewidth]{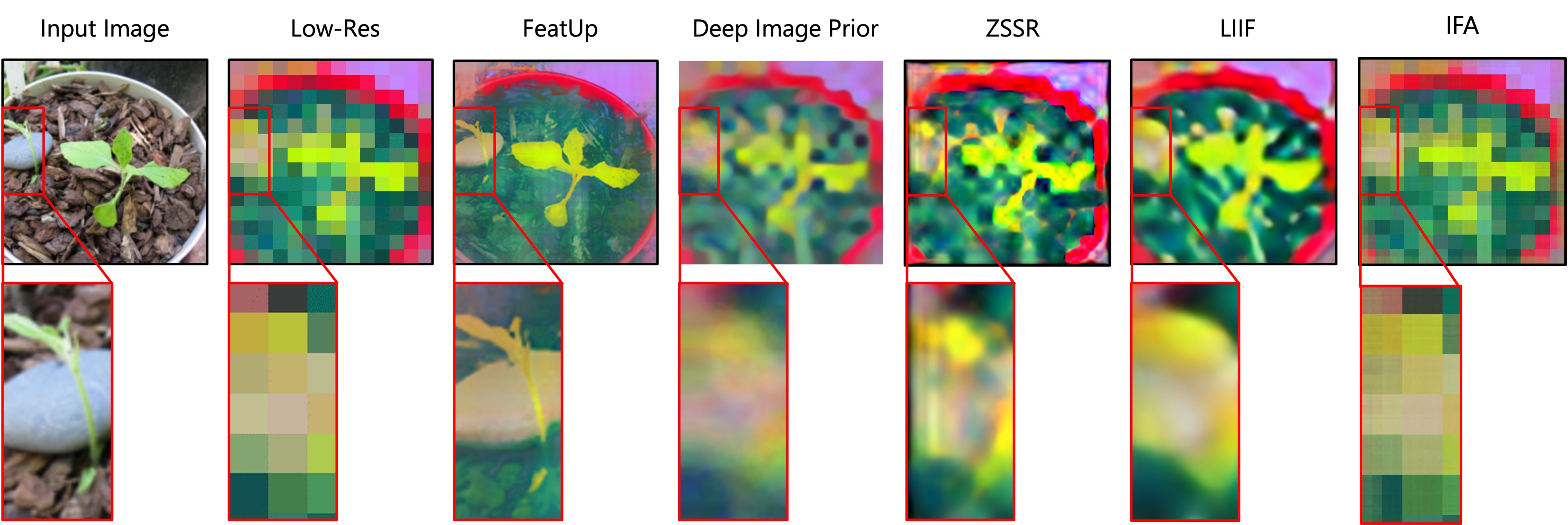}
    \caption{\small Comparison of image super-resolution methods using Deep Image Prior, Zero-Shot Super-Resolution (ZSSR), and Local Implicit Image Function (LIIF). We also include a visualization on Implicit Feature Alignment (IFA). As shown in the whole feature map and zoomed-in section, thse image upsampling methods do not effectively upsample the low-resolution and high-dimensional feature maps by the large upsampling factors that we are able to handle.}
    \label{fig:baselines}
\end{figure}

\newpage 

\subsection{Ablation Studies}
\label{subsec:ablation}
We show the effects of each design decision for FeatUp in Figure \ref{fig:ablation}. Our upsampler blurs ResNet features without the uncertainty loss, possibly because it cannot ignore certain nonlinear artifacts or resolve the large pooling window present in ResNet-50. The magnitude regularizer provides smoothing and regularization benefits. Our choice to include Fourier color features dramatically improves resolution and high-frequency details. Finally, the attention downsampler helps the system avoid odd edge and halo effects by learning kernels more focused on salient parts of the signal. Using an explicit buffer of features instead of an implicit network yields significant artifacts, though we note that the artifacts are significantly less dramatic if the simple downsampler is also used.

We also provide an ablation study of the total variation and magnitude regularizers in Figure \ref{fig:ablation-rn50}. Our regularizer is fairly robust to different settings as shown by the 2x multiplication for both terms in the 3rd column. However, there still exists an optimal $\lambda$ range that provide important smoothing properties; larger values can interfere with the main reconstruction objective as shown in the final column. 

\begin{figure}[H]
    \centering
    \includegraphics[width=\linewidth]{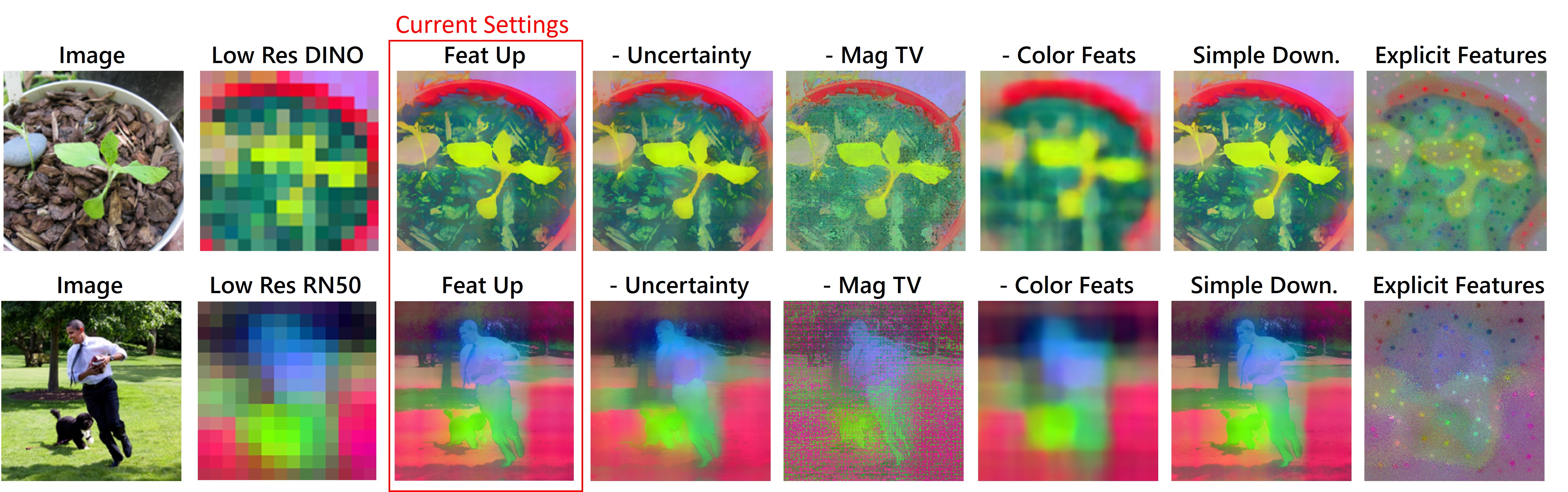}
    \caption{Qualitative ablation study across both DINO and Resnet50 Backbones. The biggest improvements arise from the implicit featurizer, color features, and the magnitude TV regularizer.}
    \label{fig:ablation}
\end{figure}

\begin{figure}[H]
    \centering
    \includegraphics[width=.7\linewidth]{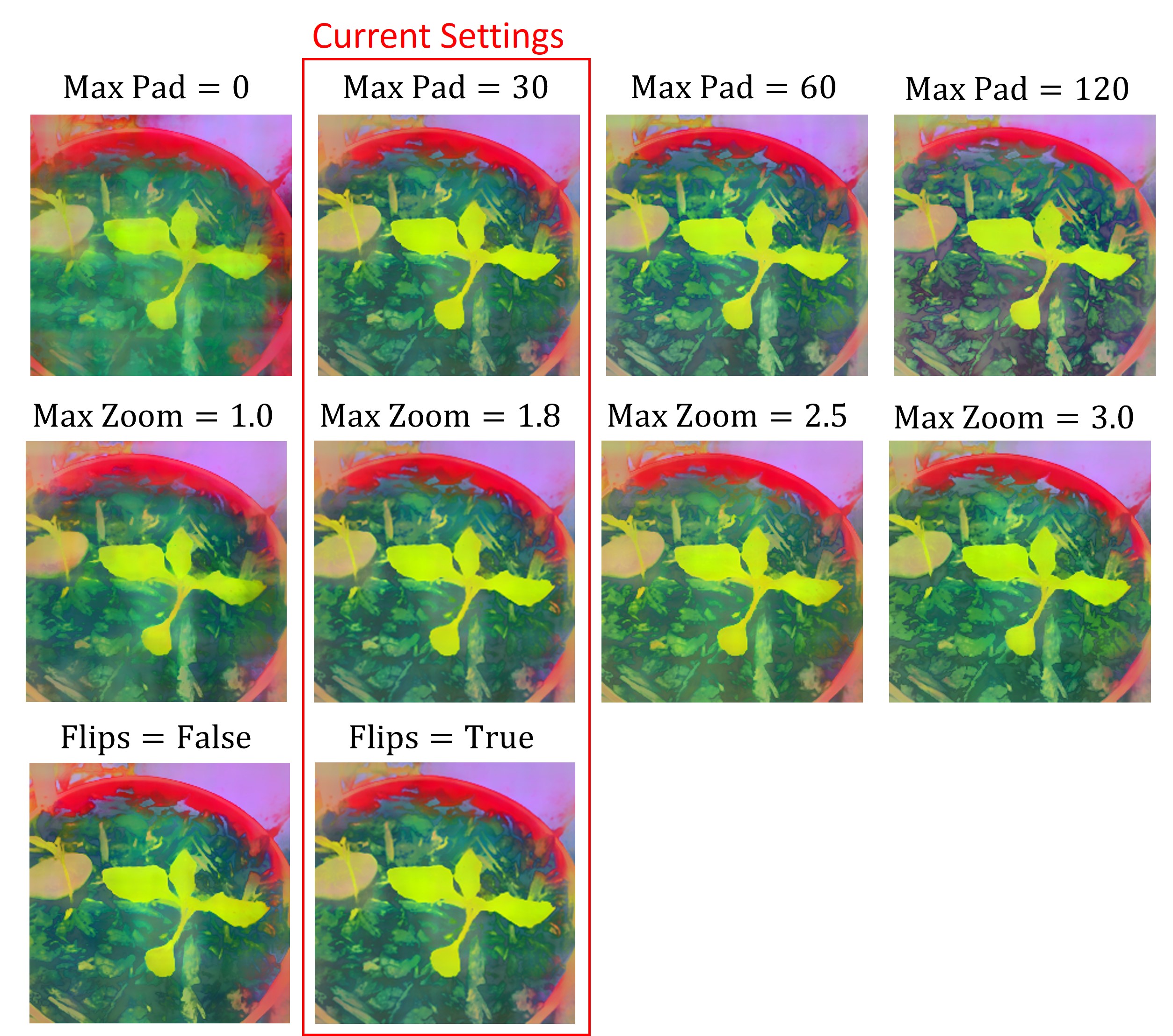}
    \caption{Ablation of FeatUp's training hyper-parameters. We are robust to a range of jitter values, though features degrade with large changes in max pad.}
    \label{fig:ablation_ranges}
\end{figure}

\begin{figure}[H]
    \centering
    \includegraphics[width=.7\linewidth]{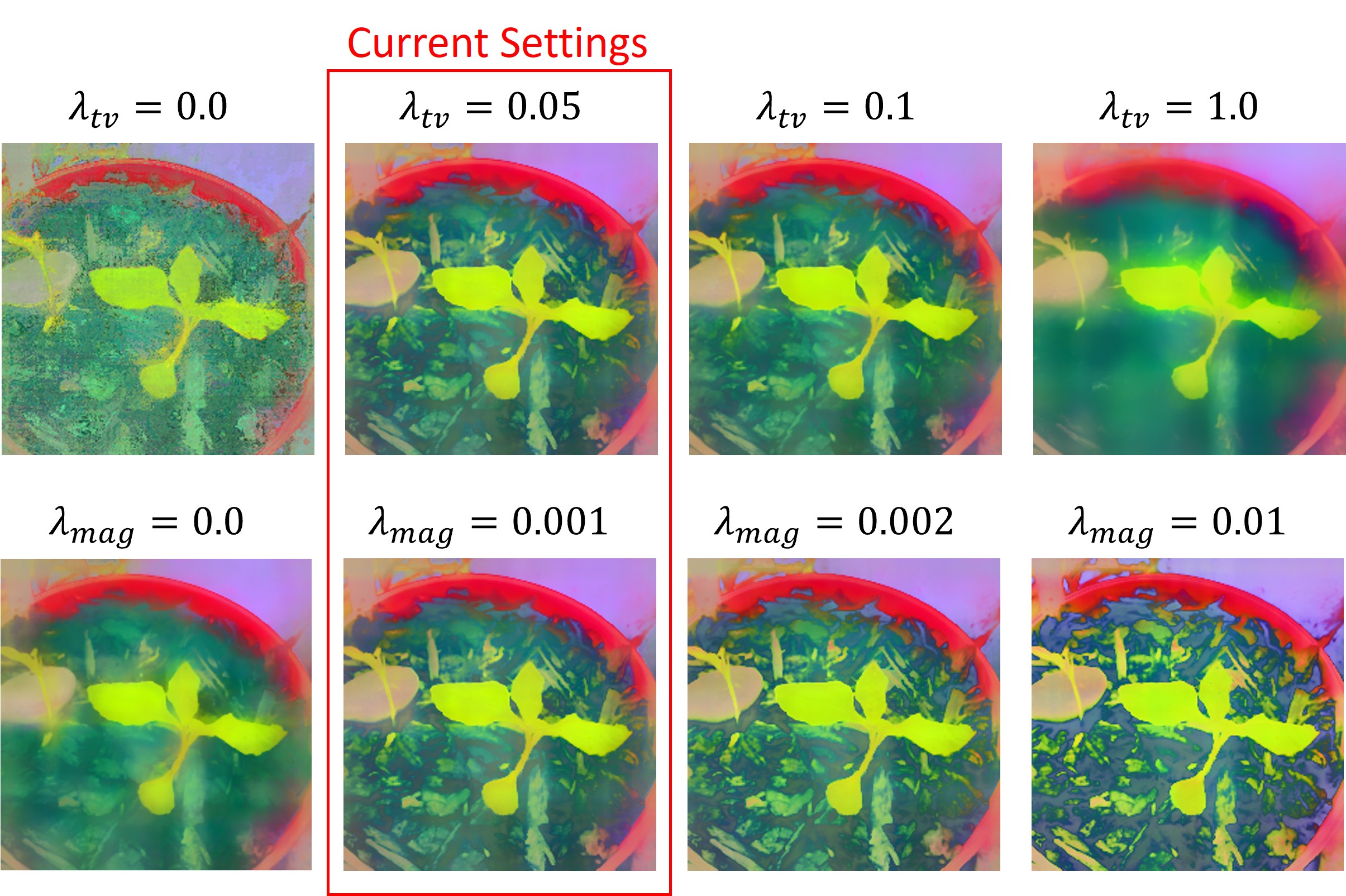}
    \caption{Qualitative ablation study of the TV and magnitude regularizers. FeatUp is fairly robust to the settng of these parameters.}
    \label{fig:ablation-rn50}
\end{figure}

To further justify our design decisions in the context of an end-to-end trained architecture, we evaluate JBU with the Segformer \cite{segformer} decoder by 1) removing the MLP (denoted as $MLP$ in Equation \ref{eq:krange}) on the guidance signal, 2) removing the temperature-weighted softmax and replacing it with Euclidean distance between the central feature and its neighborhood, and 3) removing the softmax and replacing it with cosine distance. Each ablation degrades segmentation performance, with the MLP exclusion being the most detrimental.
\begin{table}[h]
  \centering
  \renewcommand*{\arraystretch}{1.15}
\begin{tabular}{@{}ccccc@{}}
\toprule
\multicolumn{1}{l}{} & \multicolumn{4}{c}{\textbf{FeatUp (JBU)}} \\ \midrule
\multicolumn{1}{l}{\textbf{}} & Original & - MLP & \begin{tabular}[c]{@{}c@{}}- Softmax\\ + Euclidean Dist.\end{tabular} & \begin{tabular}[c]{@{}c@{}}- Softmax\\ + Cosine Dist.\end{tabular} \\ \midrule
mIoU & 44.2 & 42.9 & 43.8 & 43.7 \\
mAcc & 55.8 & 54.7 & 54.5 & 55.3 \\
aAcc & 80.7 & 79.4 & 80.0 & 80.4 \\ \bottomrule
\end{tabular}
\caption{\small Semantic segmentation performance with the Segformer architecture trained on the ADE20k train set and evaluated on the val set. Ablated FeatUp (JBU) replaces the original feature upsampling in the Segformer decoder.}
  \label{tab:jbu_ablation}
\end{table}

\begin{table}[h]
  \centering
  \renewcommand*{\arraystretch}{1.55}
\begin{tabular}{@{}rcccccc@{}}
\toprule
\multicolumn{1}{c}{} & \multicolumn{2}{c}{CAM Score} & \multicolumn{2}{c}{Semantic Seg.} & \multicolumn{2}{c}{Depth Estimation} \\
\multicolumn{1}{c}{Ablation} & A.D. $\downarrow$ & A.I. $\uparrow$ & Acc. $\uparrow$ & mIoU $\uparrow$ & RMSE $\downarrow$ & $\delta$ \textgreater 1.25 $\uparrow$ \\ \midrule
Original & \textbf{9.83} & \textbf{5.24} & \textbf{68.77} & \textbf{43.41} & \textbf{1.09} & \textbf{0.938} \\
\midrule
- MLP & 10.04 & 5.10 & 68.12 & 42.99 & 1.14 & 0.917 \\
\midrule
\begin{tabular}[c]{@{}r@{}}- Softmax\\ + Euclidean\end{tabular} & 9.98 & 5.19 & 68.68 & 43.16 & 1.10 & 0.928 \\
\midrule
\begin{tabular}[c]{@{}r@{}}- Softmax\\ + Cosine\end{tabular} & 9.97 & 5.21 & 68.49 & 43.15 & 1.12 & 0.924 \\ \bottomrule
\end{tabular}
\label{tab:jbu_ablation_2}
\caption{\small FeatUp (JBU) performance with ablated architectural components: removing the MLP, replacing softmax with a gaussian kernel w.r.t. Euclidean or cosine distance. Across all metrics, each ablation degrades performance.}
\end{table}

\begin{table}[ht]
\centering
  \renewcommand*{\arraystretch}{1.15}
\begin{tabular}{@{}ccccccccc@{}}
\toprule
\multicolumn{1}{l}{} & \multicolumn{1}{l}{} & \multicolumn{1}{l}{} & \multicolumn{2}{c}{CAM Score} & \multicolumn{2}{c}{Semantic Seg.} & \multicolumn{2}{c}{Depth Estimation} \\
Attn DS. & O.D. & TV Reg. & A.D. $\downarrow$ & A.I. $\uparrow$ & Acc. $\uparrow$ & mIoU $\uparrow$ & RMSE $\downarrow$ & $\delta$ \textgreater 1.25 $\uparrow$ \\ \midrule
\cmark & \cmark & \cmark & \textbf{8.84} & \textbf{5.60} & \textbf{71.58} & \textbf{47.37} & \textbf{1.04} & \textbf{0.927} \\
\xmark & \cmark & \cmark & 9.07 & 5.06 & 70.95 & 46.79 & 1.11 & 0.916 \\ 
\cmark & \xmark & \cmark & 8.91 & 5.55 & 71.26 & 46.89 & 1.08 & 0.920 \\
\cmark & \cmark & \xmark & 9.10 & 5.00 & 68.06 & 44.36 & 1.11 & 0.913 \\\bottomrule
\end{tabular}
\label{tab:implicit_ablation}
\caption{\small Ablation study for implicit FeatUp features with varied downsampler (attention = \cmark, simple = \xmark), outlier detection, $\lambda_{TV}$ (0.05 = \cmark, 0.0 = \xmark).}
\end{table}

\newpage

\subsection{Visualizing Additional PCA Components}
\begin{figure}[H]
    \centering
    \includegraphics[width=0.8\linewidth]{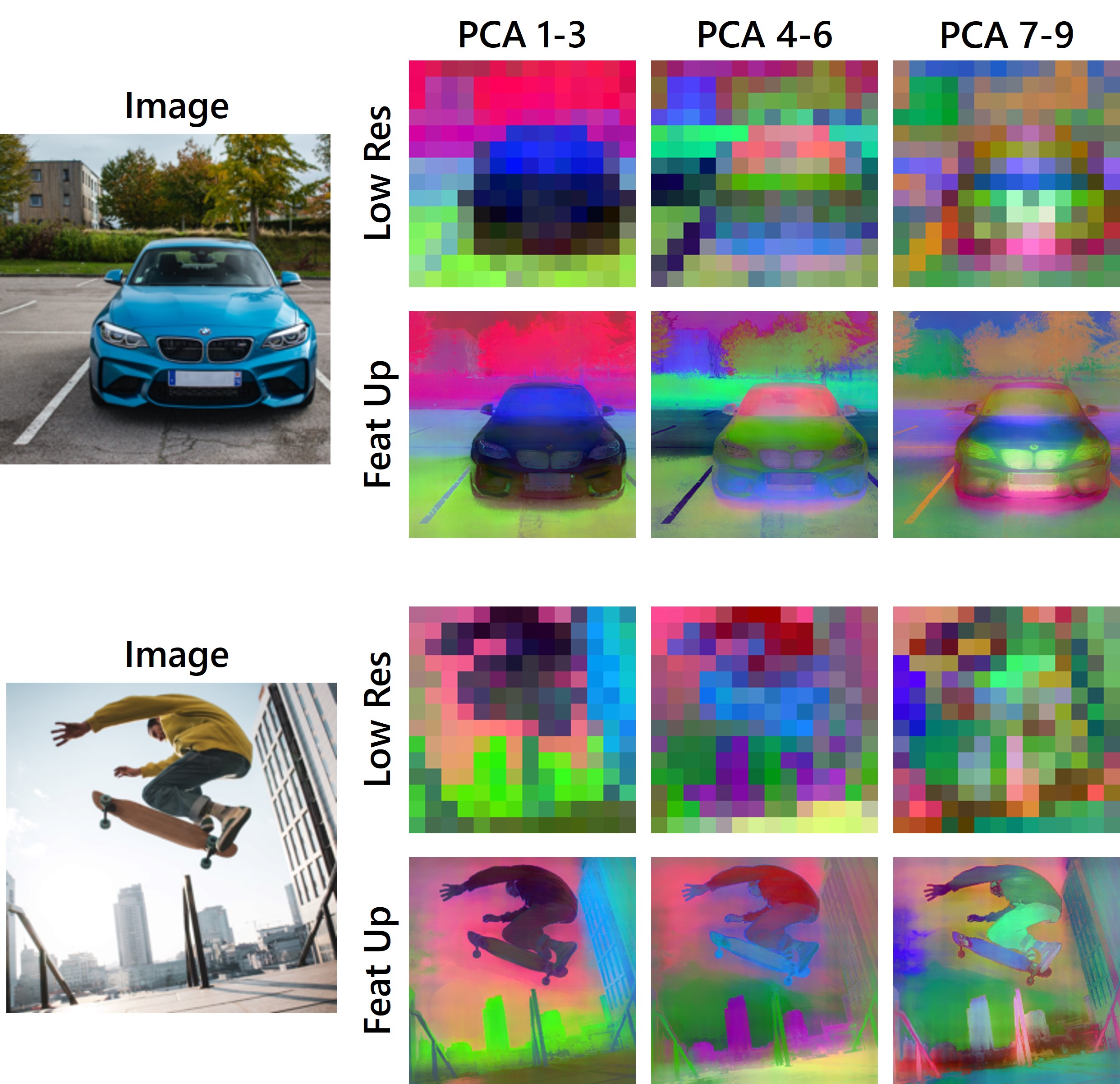}
    \caption{Visualizing higher PCA components with FeatUp. FeatUp upsamples entire feature maps, so their higher-order principal components also remain in the same space as the original features and are upsampled precisely. Higher components tend to separate more fine-grained object categories like the skater from the skateboard, and the trees from the background, and the clouds from the sky. Note that each subobject's features are upsampled precisely to the object it represents.}
    \label{fig:pca_viz}
\end{figure}
\newpage
\subsection{Saliency Map Details}
Downsampling in FeatUp is analogous to ray-marching in NeRF, which approximates the physics of image formation. FeatUp’s downsampler approximates a network’s process of pooling information into features. As shown in Figure 6, most networks preserve the rough location of objects in their features (the objects just appear downsampled and blurred). This observation leads us to use blur/pooling operators.

The simplest of these is average pooling, but we can do better by generalizing this operation to a learned blur/pooling kernel so the downsampler can better match a network’s receptive field size. To map back to NeRF, this is like adding learned camera lens distortion parameters to the ray-marcher so NeRF can better fit the data.

As shown in Figure \ref{fig:backbone_comp} and described in Section 3.1, even a learned blur/pooling kernel cannot capture dynamic receptive fields or object salience. For example if a small amount of an important object is in a transformer’s patch, the whole feature changes. We capture effects like this by making the learned pool/blur kernel dependent on image content using a 1x1 conv (we don’t need anything bigger than this layer). This generalizes the previously-described learned blur/pool and allows the downsampler to adaptively pool based on image content. Figure \ref{fig:downsampler-attention} shows that the salience network focuses on certain attributes (e.g. object boundaries, some important small objects). We also note that many common pooling strategies such as average pooling or nearest/bilinear/bicubic resizing are special cases of our learnable attention pooling strategy.

\subsection{Visualizing Downsampler Salience and Kernels}

\begin{figure}[H]
    \centering
    \includegraphics[width=0.7\linewidth]{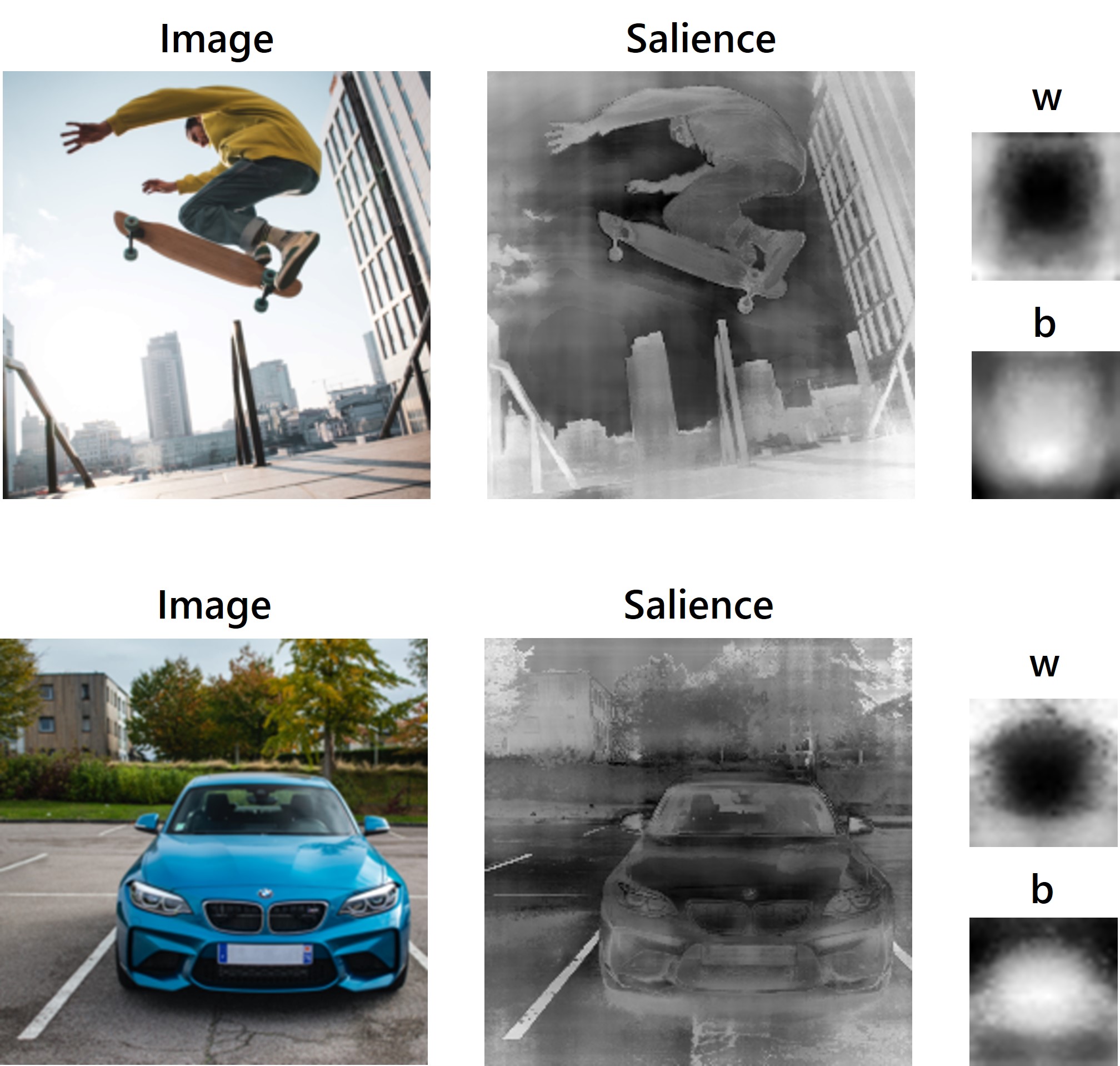}
    \caption{Visualization of downsampler salience and weight and bias kernels for two images. Note how fine-grained objects have higher salience and regions around important objects (like the sky between the hands and the skateboard) have lower salience. This allows the network to capture nonlinear behavior where embeddings from salient regions dominate the embeddings of other regions.}
    \label{fig:downsampler-attention}
\end{figure}

\subsection{Visualizing Predicted Uncertainty}
\begin{figure}[H]
    \centering
    \includegraphics[width=0.7\linewidth]{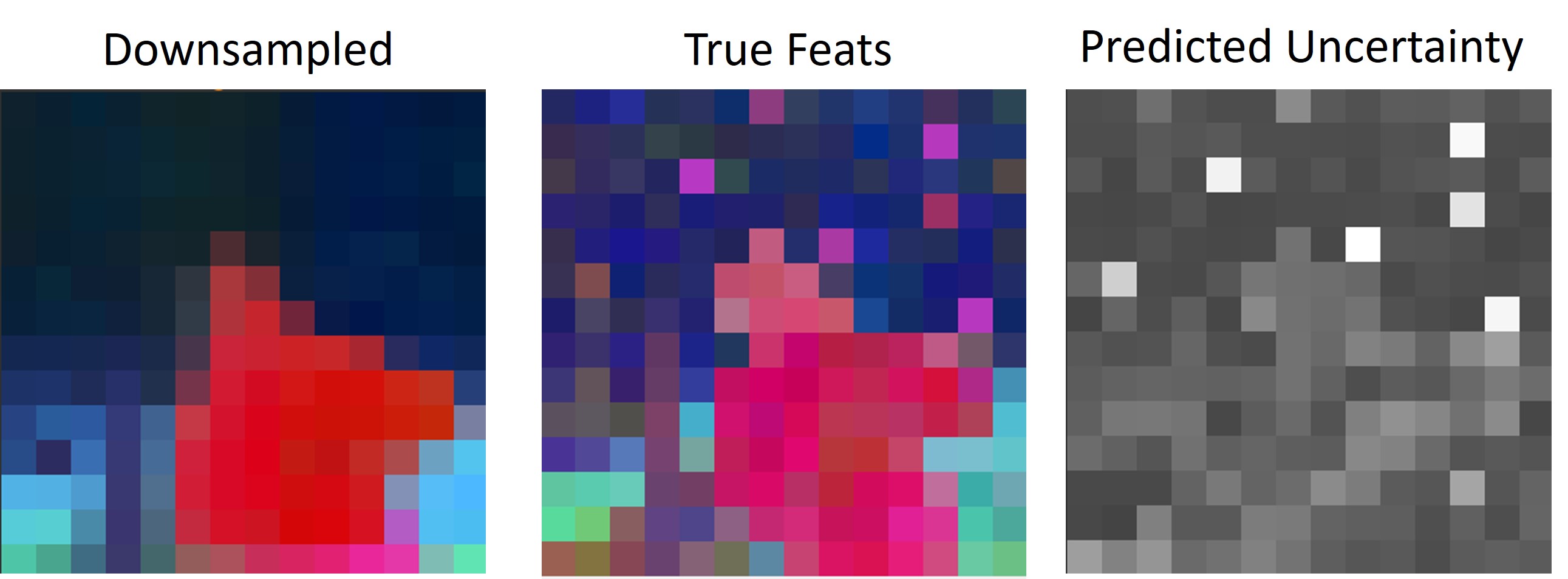}
    \caption{An example predicted uncertainty map for a set of ViT features. White areas have higher uncertainty. In this figure, we can see that nonlinear artifacts like the spurious pink tokens are marked with high uncertainty as they change location depending on the given evaluation. These tokens might serve some other role in the network, such as class-token-like information aggregation. We do not see these types of effects in DINO or convolutional networks. }
    \label{fig:predicted_uncertainty}
\end{figure}

\subsection{Improving Image Retrieval for Small Objects}

\begin{figure}[H]
    \centering
    \includegraphics[width=0.9\linewidth]{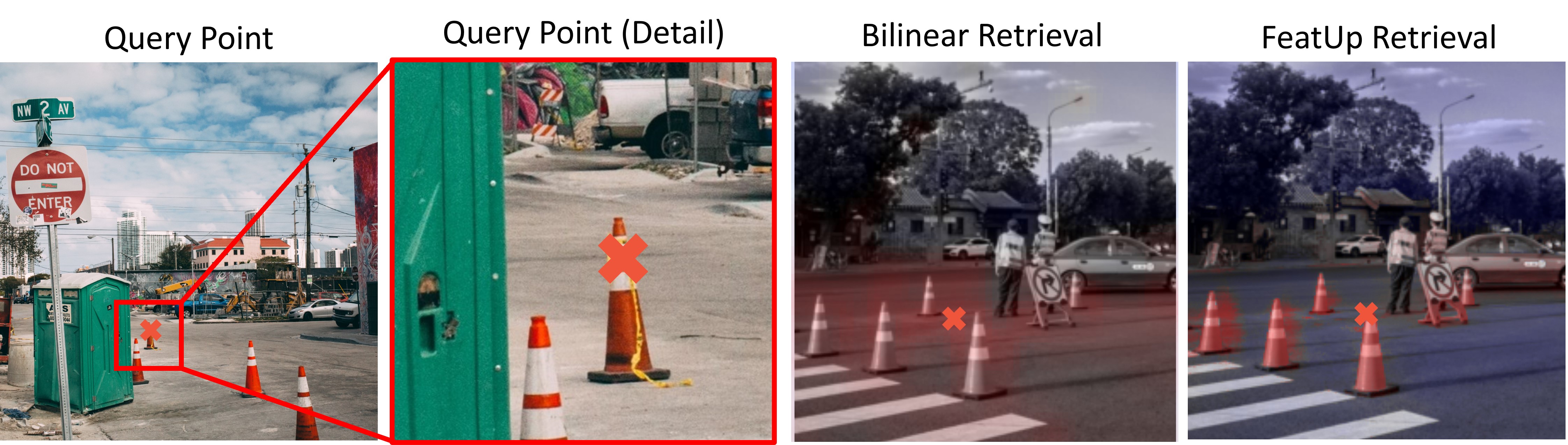}
    \caption{\label{fig:supp_retrieval}
    High-resolution FeatUp features can be used to improve the retrieval of small objects and cluttered scenes. A query image (Left) is featurized with DINO and the region marked with a red $\times$ is used as a query point. We show the detailed placement of this query point in the second image from the left. In the two images on the right, we show the closest matching point in the target image (red $\times$) and we also visualize the similarity heatmap (red means similarity, blue means dissimilarity). The second image from the right depicts the matching point and heatmap when using bilinear feature interpolation on the image and target. The image on the far right shows the results after upsampling with FeatUp prior to computing the retrieval. Because the scene is cluttered, bilinear interpolation blurs object features together and the resulting query vector attends over both the ground and the traffic cones. FeatUp's features better align with objects allowing only the traffic cones to be retrieved. }
\end{figure}
\newpage

\subsubsection{Linear Probe details}
In both linear probe tasks, one probe was trained on low-resolution (14x14) features from the COCO training set, and frozen for validation across all methods. FeatUp's performance improvements on this repurposed linear probe show that our methods increase resolution without compromising the original feature space. We highlight that these results are \textit{not meant to improve state-of-the-art performance} on segmentation and depth estimation; they are meant to showcase \textit{feature quality} across upsamplers. Because prediction for both tasks is done with a frozen backbone and a single trainable linear probe, the segmentation and depth maps are not meant as a direct application. 

\subsection{Average Drop and Average Increase Details}
\label{subsec:adai}
Average Drop is expressed as $\sum\limits_{i=1}^N \frac{max(0, Y^c_i - O^c_i)}{Y_i^c} \cdot 100$, where $Y_c^i$ is the classifier's softmax output (i.e. confidence) on sample $i$ for class $c$, and $O^c_i$ is the classifier's softmax output on the CAM-masked sample $i$ for class $c$. We generate $O^c_i$ by keeping the top 50\% of CAM values (and Gaussian blurring the remaining 50\% of values with less explainability power). Though we generally expect classifiers to drop in confidence because even masking out less-salient pixels can remove important image context, a high-quality CAM will target the explainable regions of an image more precisely and thus maintain a higher confidence. In the reverse direction, we measure the Average Increase to capture the instances where CAM-masked inputs increase model confidence. Specifically, we define Average Increase as $\sum\limits_{i=1}^N \frac{\mathbbm{1}_{Y^c_i < O^c_i}}{N} \cdot 100$ where $\mathbbm{1}_{Y^c_i < O^c_i}$ is an indicator function equal to 1 when $Y^c_i < O^c_i$ - that is, when model confidence increases upon classifying a CAM-masked image.

Similar to the RelevanceCAM evaluation in \citep{relevancecam}, we randomly select 2000 images from the ImageNet validation set (limited to images where the label and model prediction match) to measure A.D. and A.I. on. 

\newpage
\subsection{Performance Benchmarking}

See Table \ref{tab:cuda_benchmark_full} for performance benchmarking of our adaptive convolution CUDA kernel used in FeatUp (JBU).

\begin{table*}[ht]
\begin{center}
\begin{tabular}{ccccc}
\toprule
Shape (B, H, W, C, F)                                                                     & Method      & Forward (ms)  & Backward (ms) & Peak Mem (Mb) \\ \midrule
\multicolumn{1}{c|}{\multirow{3}{*}{$1 	\times 14   	\times 14 	\times 2048 	\times 5$}}  & Ours        & \textbf{0.15} & \textbf{1.05} & \textbf{6.24} \\
\multicolumn{1}{c|}{}                                                                     & TorchScript & 2455          & 69367         & 12.8          \\
\multicolumn{1}{c|}{}                                                                     & Unfold      & 3.30          & 2.81          & 119.          \\ \midrule
\multicolumn{1}{c|}{\multirow{3}{*}{$1 	\times 512   	\times 512 	\times 3 	\times 5$}}   & Ours        & \textbf{0.55} & \textbf{2.10} & \textbf{10.2} \\
\multicolumn{1}{c|}{}                                                                     & TorchScript & 147.          & 520.          & 24.3          \\
\multicolumn{1}{c|}{}                                                                     & Unfold      & 3.47          & 4.85          & 231.          \\ \midrule
\multicolumn{1}{c|}{\multirow{2}{*}{$16 	\times 32   	\times 32 	\times 2048 	\times 5$}} & Ours        & \textbf{8.43} & \textbf{90.8} & \textbf{372.} \\
\multicolumn{1}{c|}{}                                                                     & Unfold      & 118.          & 218.          & 6628.         \\ \midrule
\multicolumn{1}{c|}{\multirow{2}{*}{$32 	\times 512   	\times 512 	\times 3 	\times 5$}}  & Ours        & \textbf{17.7} & 114.          & \textbf{326.} \\
\multicolumn{1}{c|}{}                                                                     & Unfold      & 36.0          & \textbf{104.} & 4901.         \\ \midrule
\multicolumn{1}{c|}{\multirow{2}{*}{$64 	\times 14   	\times 14 	\times 2048 	\times 5$}} & Ours        & \textbf{6.12} & \textbf{61.1} & \textbf{400.} \\
\multicolumn{1}{c|}{}                                                                     & Unfold      & 57.5          & 170.          & 5174.         \\ \midrule
\multicolumn{1}{c|}{\multirow{2}{*}{$64 	\times 224   	\times 224 	\times 3 	\times 5$}}  & Ours        & \textbf{6.27} & 36.1          & \textbf{128.} \\
\multicolumn{1}{c|}{}                                                                     & Unfold      & 16.7          & \textbf{27.4} & 1878.         \\ \midrule
\multicolumn{1}{c|}{\multirow{2}{*}{$64 	\times 64   	\times 64 	\times 16 	\times 5$}}   & Ours        & \textbf{1.06} & \textbf{8.99} & \textbf{44.5} \\
\multicolumn{1}{c|}{}                                                                     & Unfold      & 7.18          & 14.5          & 822.          \\ \midrule
\multicolumn{1}{c|}{\multirow{2}{*}{$64 	\times 64   	\times 64 	\times 16 	\times 7$}}   & Ours        & \textbf{2.00} & \textbf{8.36} & \textbf{52.6} \\
\multicolumn{1}{c|}{}                                                                     & Unfold      & 10.8          & 25.6          & 1596.         \\
\bottomrule
\end{tabular}
\caption{Comparing the performance of our CUDA JBU kernel with with implementations based on PyTorch's \texttt{Unfold} operation and TorchScript. Our implementation dramatically reduces memory overhead and increases inference speed. Code for this operation is available in the provided link.}
\label{tab:cuda_benchmark_full}
\end{center}
\end{table*}

\begin{figure}[H]
\centering
\vspace{-6mm}
    \includegraphics[width=\linewidth]{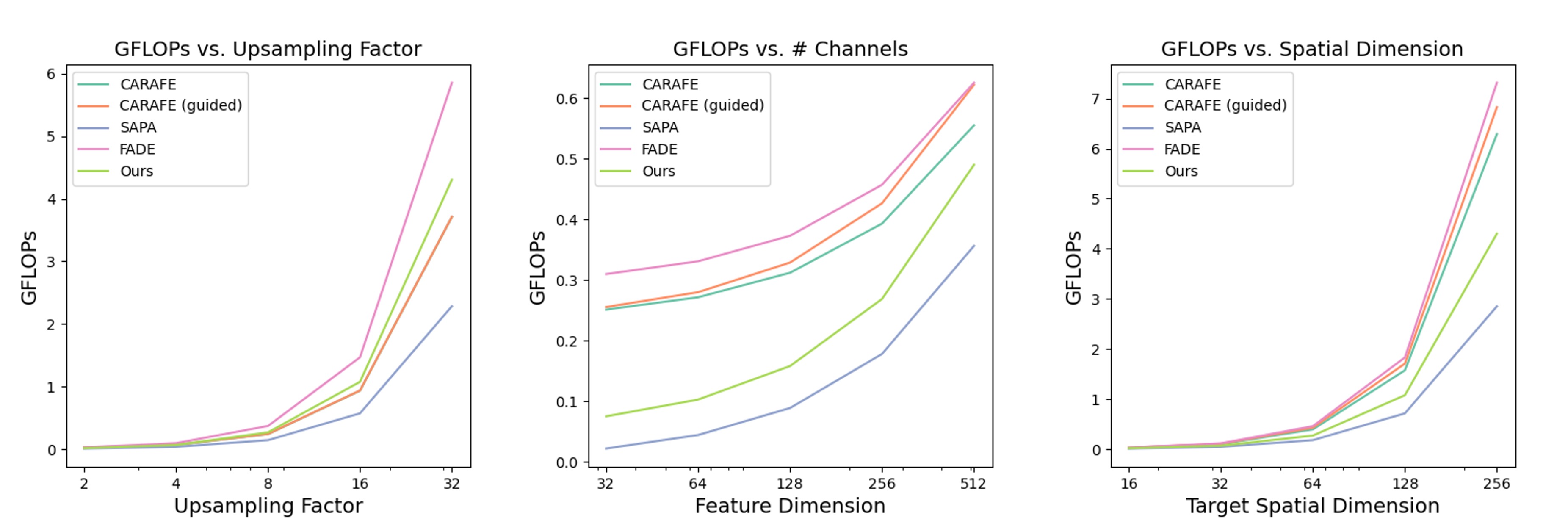}
    \caption{\small We evaluate how floating point operations scale with various factors. In varying the upsampling factor, feature dimension, and target spatial dimension, FeatUp (JBU) remains competitive in GFLOP usage. For each experiment, the attributes not studied are kept constant (upsampling factor = 2, feature dimension = 256, starting spatial dimension = 8x8).}
    \label{fig:gflops}
\end{figure}

We analyze peak memory usage and inference time for various upsampling methods. Specifically, we upsample ViT features from a $(1 \times 3 \times 224 \times 224)$ image (i.e. low-resolution feature dimensions of $(1 \times 384 \times 14 \times 14)$) by factors of 2, 4, 8, and 16. Figure \ref{fig:memory_time} shows that FeatUp (JBU)'s peak memory closely follows resize-conv and SAPA baselines and outperforms CARAFE. Additionally, FeatUp is as fast as baselines yet outperforms baselines in all our quantitative evaluations. We note that strided and large image baselines become computationally infeasible after $8\times$ upsampling, even using a batch size of 1.

\begin{figure*}[h]
    \centering
    \includegraphics[width=\linewidth]{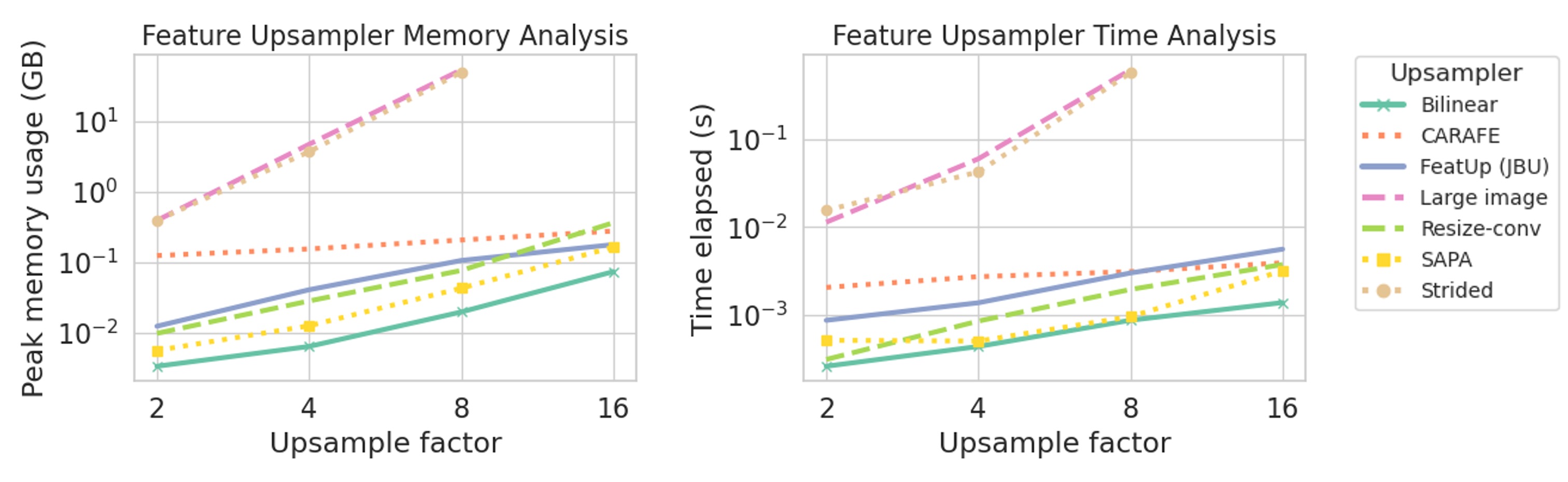}    
    \vspace{-.1in}
    \caption{\small Analysis of peak memory usage (left) and inference time (right) for various forward-pass upsamplers. FeatUp (JBU) is competitive with SAPA and resize-conv across upsampling factors and is more efficient than CARAFE for smaller factors. The large image and strided approaches become infeasible at large upsampling factors we only show metrics for these methods up to $8\times$ upsampling.}
    \vspace{-.15in}
    \label{fig:memory_time}
\end{figure*}
\newpage

\subsection{Additional Qualitative Results}
We provide additional CAM visualizations with supervised ViT features on the ImageNet val set in Figure \ref{fig:supp_cam_vit}. As in the main paper, we upsample features from 14x14 to 224x224 output before extracting CAMs (except for the ``Low-Res" column, where the features are kept as-is). Both FeatUp (JBU)'s edge-preserving bilateral filters and the FeatUp (Implicit)'s feature representation allow resulting CAMs to highlight salient regions more accurately. Our CAMs combine the semantic advantages of low-resolution features and the spatial advantages of large images, producing refined versions of the original CAMs without discontinuous patches present in the other upsampling schemes.

\begin{figure}[H]
    \centering
    \includegraphics[width=1\linewidth]{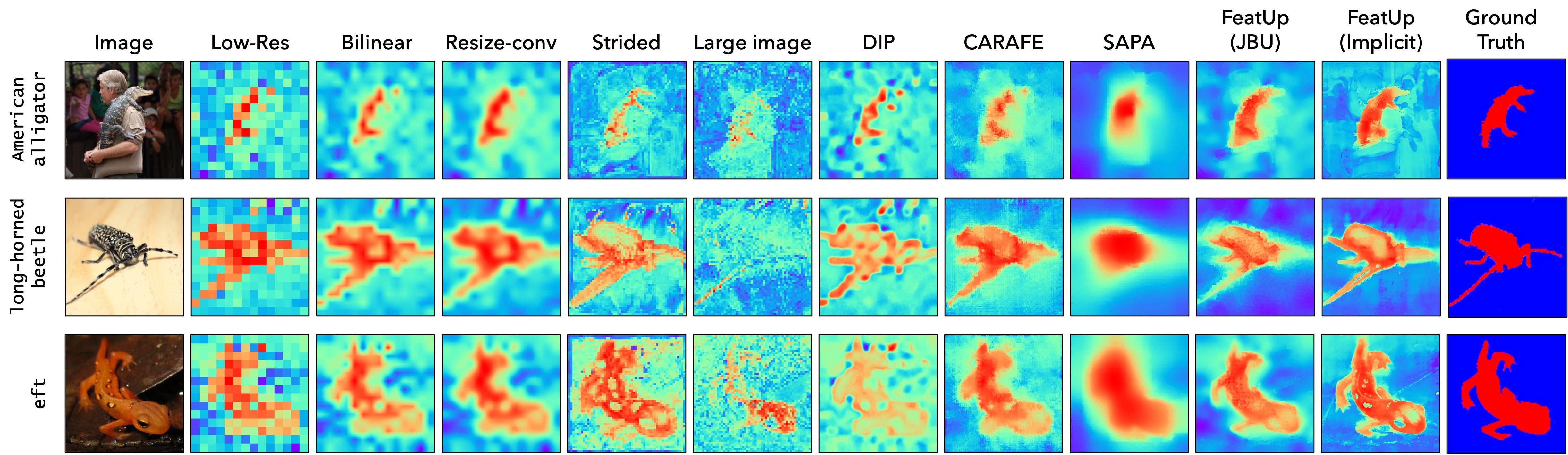}
    \caption{\label{fig:supp_cam_vit}\small 
    CAMs on the ImageNet validation set from a supervised ViT backbone and linear probe classifier. Both FeatUp variants produce features that are more precise with respect to the input image, allowing downstream CAMs to better align with object boundaries. }
\end{figure}

See Figure \ref{fig:supp_seg} for examples of linear probe transfer learning for semantic segmentation on the COCO-Stuff dataset. The 14x14 features output from a ViT backbone are upsampled with the following methods to achieve 224x224 resolution. Then, a linear probe is trained on the low-resolution features and frozen for evaluation on COCO-Stuff semantic class labels.  Our methods recover more cohesive labels of objects and backgrounds.

\begin{figure}[H]
    \centering
    \includegraphics[width=1\linewidth]{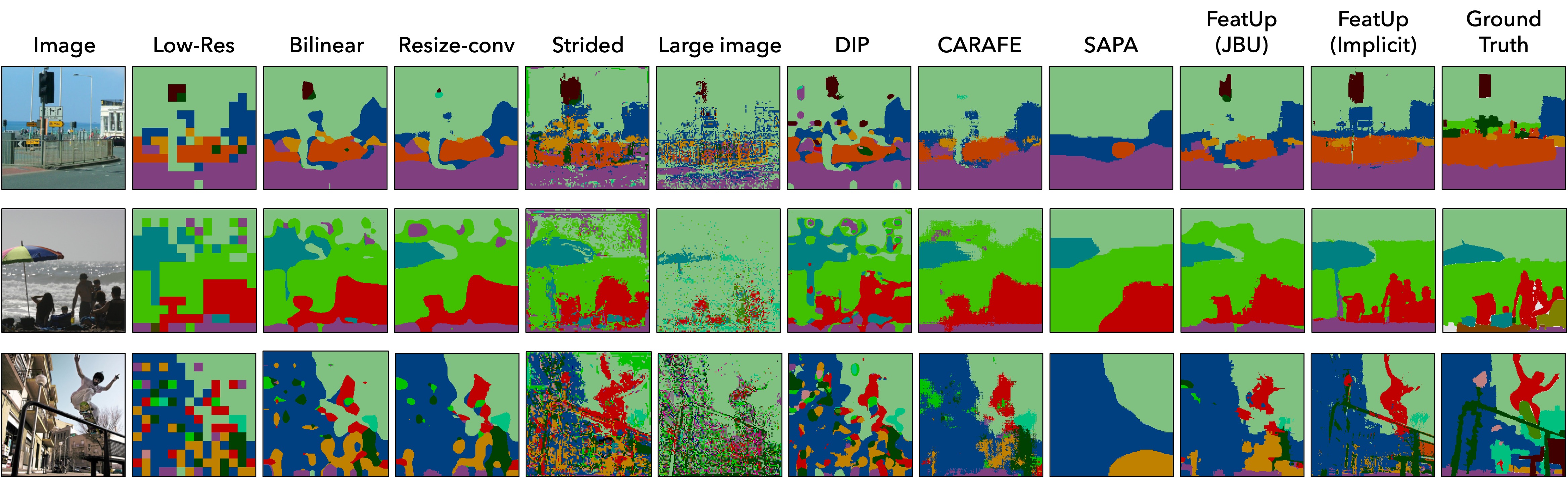}
    \caption{\label{fig:supp_seg}\small 
    Examples of linear probe transfer learning for semantic segmentation on the COCO-Stuff dataset. Our methods more closely resemble ground-truth segmentation and smooth many of the artifacts present in the low-resolution features. Additionally, FeatUp (Implicit) recovers thin structures like the umbrella pole not even present in the ground truth despite being semantically correct.}
\end{figure}

Figure \ref{fig:supp_depth} provides additional examples of linear probe transfer learning for depth estimation. The 14x14 features output from a ViT backbone are upsampled to achieve 224x224 resolution. Then, a linear probe is trained \textit{directly on the features} to predict depth while supervised with a small MiDaS network. Our results show that both FeatUp variants result in high-quality features capable of transfer learning.

\begin{figure}[H]
    \centering
    \includegraphics[width=1\linewidth]{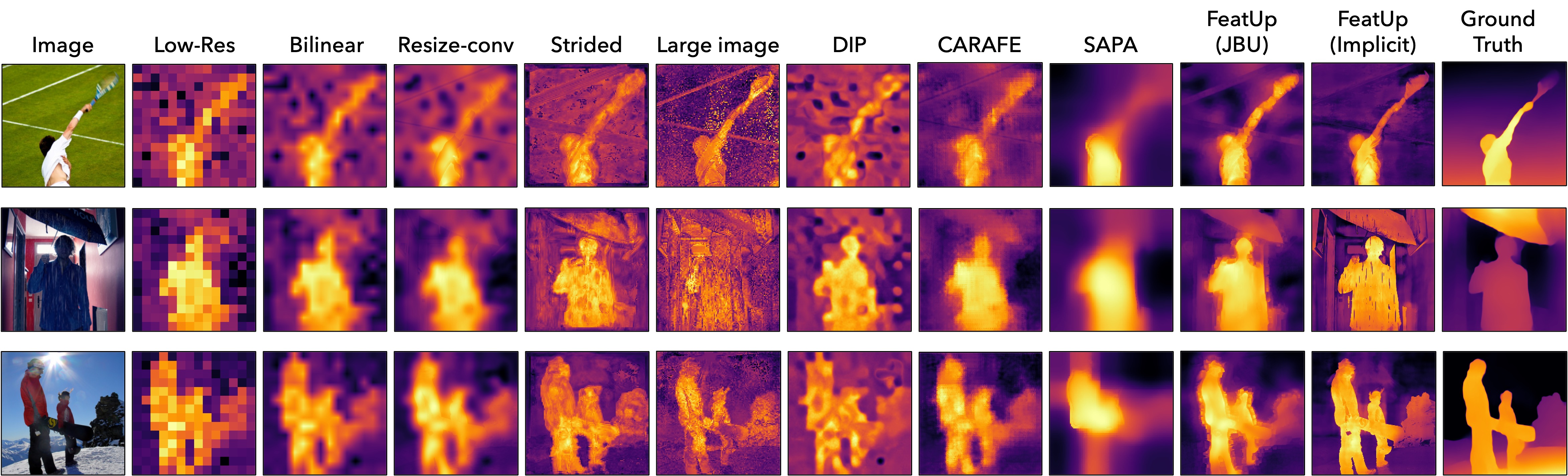}
    \caption{\label{fig:supp_depth}\small 
    Examples of linear probe transfer learning for depth estimation. Our methods produce sharper object boundaries and smoother interiors that more closely align with true depth than other methods. }
\end{figure}

\begin{figure*}[h]
    \centering
    \includegraphics[width=\linewidth]{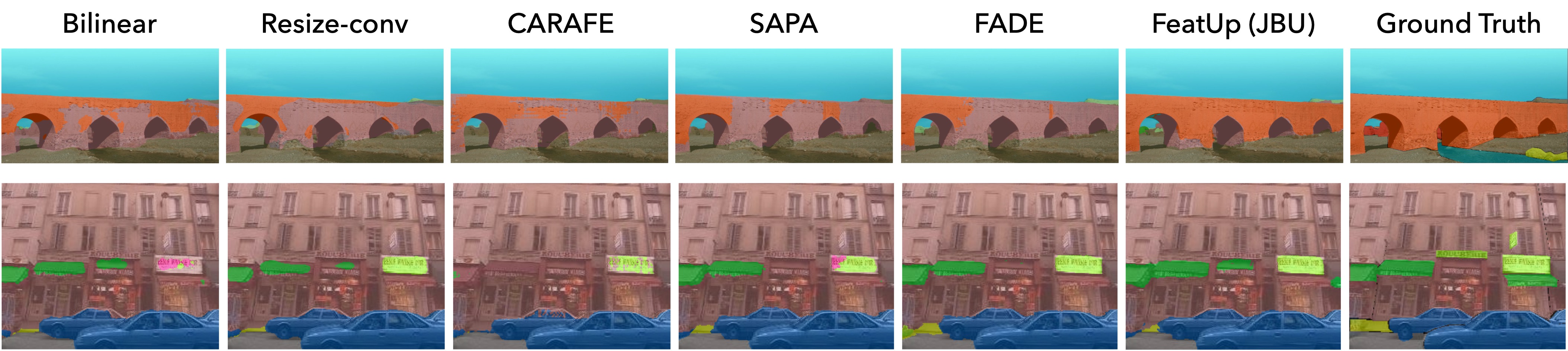}    
    \vspace{-.2in}
    \caption{\small End-to-end training performance of different upsampling methods from our Segformer based semantic segmentation experiments. These results do not use linear probes, but instead train the architecture jointly.}
    \label{fig:seg_viz}
\end{figure*}

\newpage

\subsection{Limitations}

\begin{figure}[H]
    \centering
    \includegraphics[width=0.6\linewidth]{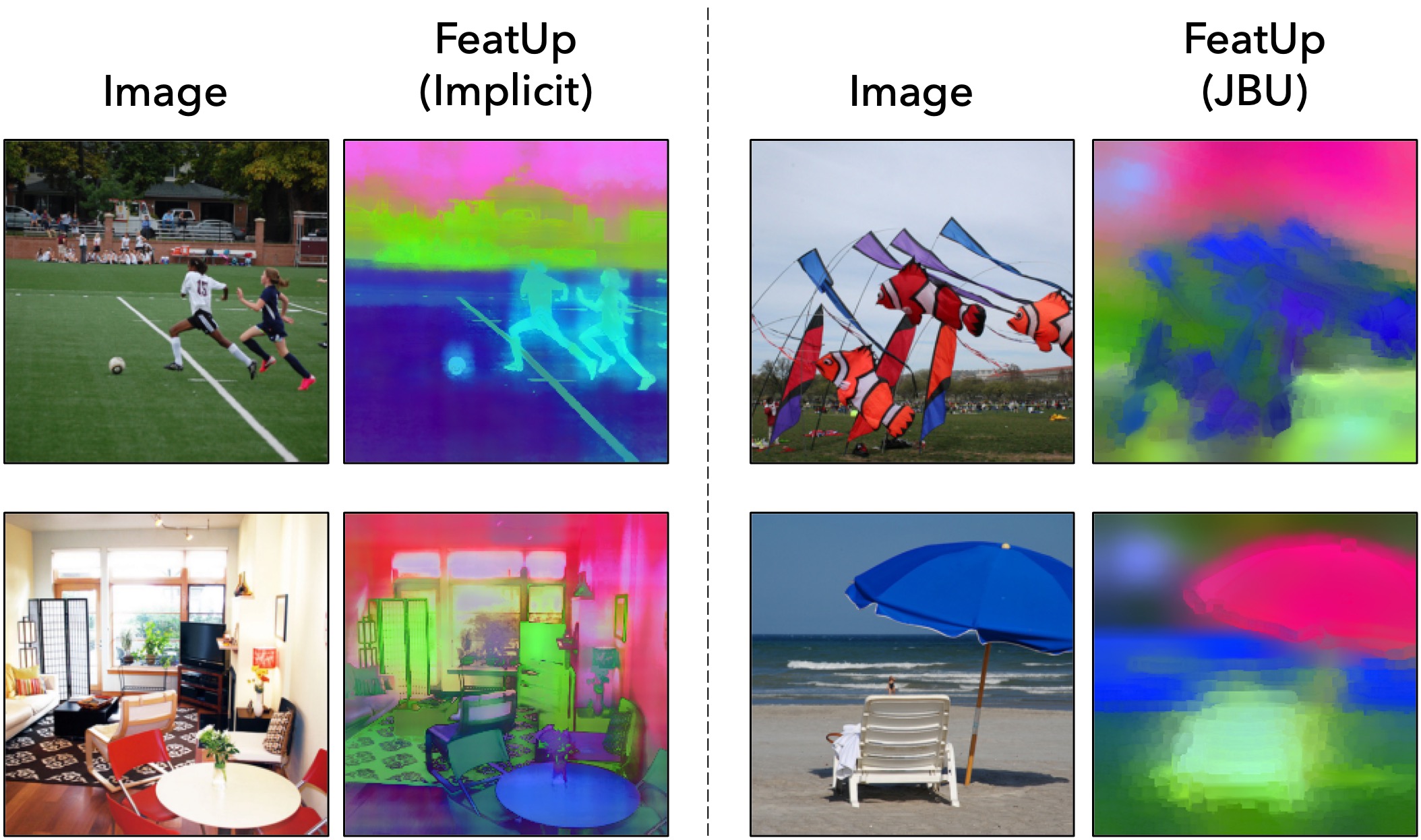}
    \caption{\label{fig:limitations}\
    Left: Though FeatUp's implicit network can capture fine detail such as the soccer ball or window frame, it can still produce some halo effects (see soccer player). Additionally, because the method relies on the input image's spatial signal, certain patterns unrelated to object semantics can be transferred to the feature map (see rug pattern), though this is a rare occurrence. Right: FeatUp's JBU network is not as sensitive to fine detail as the implicit network, instead capturing broader contours.}
\end{figure}

\subsection{Implementation Details}
All backbones (DINO, DINOv2, ViT, ResNet-50, CLIP, and DeepLabV3) used to train FeatUp are frozen, pre-trained models obtained from the community. We outline the hyperparameters used to train FeatUp in table \ref{tab:featup-hyperparams}.

\begin{table}[H]
\centering
\begin{tabular}{c|cc}
\toprule
\textbf{Hyperparameter} & \textbf{FeatUp (Implicit)} & \textbf{FeatUp (JBU)} \\ \midrule
Num Images                 & 1               & 4                   \\
Num Jitters Per Image      & 10              & 2                   \\
Downsampler                & Attention       & Attention           \\
Optimizer                  & NAdam           & NAdam               \\
Learning Rate              & 0.001           & 0.001               \\
Image Load Size            & 224             & 224                 \\
Projection Dim             & 128             & 30                  \\
Training Steps             & 2000            & 2000              \\
Max Transform Padding      & 30px            & 30px                \\
Max Transform Zoom         & 1.8$\times$     & 2$\times$           \\
Kernel Size                & 29              & 16                  \\
Total Variation Weight     & 0.05            & 0.0                 \\
Implicit Net Layers        & 3               & n/a                 \\
Implicit Net Dropout       & 0.1             & n/a                 \\
Implicit Net Activation    & ReLU            & n/a                 \\ \bottomrule
\end{tabular}
\vspace{.1in}
\caption{Hyperparameters used in training FeatUp.}
\label{tab:featup-hyperparams}.
\end{table}

\end{document}